\documentclass[journal,onecolumn]{IEEEtran}

\usepackage{times}
\usepackage{epsfig}
\usepackage{graphicx}
\usepackage{amsmath}
\usepackage{amssymb}
\usepackage{algorithmic}
\usepackage[linesnumbered,ruled,vlined]{algorithm2e}
\usepackage{multirow}
\usepackage{booktabs}
\usepackage{siunitx}
\usepackage{bm}
\usepackage{amsfonts}
\usepackage{caption}

\usepackage{CJK}
\newif\ifdraft\drafttrue
\ifdraft

\fi
\usepackage[pagebackref=true,breaklinks=true,letterpaper=true,colorlinks,bookmarks=false]{hyperref}

\ifCLASSINFOpdf

\else

\fi

\begin{document}
\begin{CJK*}{UTF8}{gkai}
\title{ Discriminative and  Geometry Aware Unsupervised Domain Adaptation}

\author{Lingkun Luo, ~Liming~Chen,~\IEEEmembership{Senior~member,~IEEE,  Shiqiang Hu,  Ying Lu and, Xiaofang Wang,}
         
\thanks{K. Luo, S. Qiang are with School of Aeronautics and Astronautics, Shanghai Jiao Tong University, 800 Dongchuan Road, Shanghai, China e-mail: lolinkun@gmail.com, sqhu@sjtu.edu.cn.}
\thanks{K. Luo, L. Chen, Y. Lv and  X. Wang are with LIRIS, CNRS UMR 5205, 	Ecole Centrale de Lyon, 36 avenue Guy de Collongue, Ecully,  France e-mail: (liming.chen,ying.lu,xiaofang.wang, )@ec-lyon.fr.}
\thanks{Manuscript received December 27, 2017.}}

\markboth{Journal of \LaTeX\ 2017}%
{Shell \MakeLowercase{\textit{et al.}}: Bare Demo of IEEEtran.cls for IEEE Journals}

\maketitle

\begin{abstract}
Domain adaptation (DA)  aims to  generalize a learning model across training and testing data despite the mismatch of their data  distributions. In light of a theoretical estimation of upper error bound, we argue in this paper that an effective DA method should 1) search a shared feature subspace where source and target data are not only aligned in terms of distributions as most state of the art DA methods do, but also discriminative in that instances of different classes are well separated; 2) account for the  geometric structure of the underlying data manifold when inferring data labels on the target domain. In comparison with a baseline DA method which only cares about data distribution alignment between source and target, we derive three different DA models, namely \textbf{CDDA}, \textbf{GA-DA}, and \textbf{DGA-DA}, to highlight the contribution of Close yet Discriminative DA(CDDA) based on 1), Geometry Aware DA (\textbf{GA-DA}) based on 2), and finally Discriminative and Geometry Aware DA (DGA-DA) implementing jointly 1) and 2). Using both synthetic and real data, we show the effectiveness of the proposed approach which consistently outperforms state of the art DA methods over 36 image classification DA tasks through 6 popular benchmarks. We further carry out in-depth analysis of the proposed DA method in quantifying the contribution of each term of our DA model and provide insights into the proposed DA methods in visualizing both real and synthetic data.  

\end{abstract}


%
\IEEEpeerreviewmaketitle

\section{Introduction}

\IEEEPARstart{T}{raditional} machine learning tasks assume that both training and testing data are drawn from a same data distribution\cite{pan2010survey,7078994,DBLP:journals/corr/Csurka17}. However, in many real-life applications, due to different factors as diverse as sensor difference, lighting changes, viewpoint variations, \textit{etc.}, data from a target domain may have a different data distribution \textit{w.r.t.} the labeled data in a source domain where a predictor can be  can not be reliably learned due to the data distribution shift. On the other hand, manually labeling enough target data for the purpose of training an effective predictor can be very expensive, tedious and thus prohibitive.  
    
    Domain adaptation (DA) \cite{pan2010survey,7078994,DBLP:journals/corr/Csurka17} aims to leverage possibly abundant labeled data from a \textit{source} domain to learn an effective predictor for data in a \textit{target} domain despite the data distribution discrepancy between the source and target. While DA can be \textit{semi-supervised} by assuming a certain amount of labeled data is available in the target domain, in this paper we are interested in \textit{unsupervised} DA\cite{DBLP:conf/icml/SaitoUH17} where we assume that the target domain has no labels. 
    

State of the art DA methods can be categorized into \textit{instance}-based \cite{pan2010survey,donahue2013semi}, \textit{feature}-based \cite{Busto_2017_ICCV,long2013transfer,DBLP:journals/tip/XuFWLZ16}, or \textit{classifier}-based. Classifier-based DA is not suitable to unsupervised DA as it aims to fit a classifier trained on the source data to the target data through adaptation of its parameters, and thereby require some labels in the target domain\cite{tang2017visual} .  The instance-based approach generally assumes that   1) the conditional distributions of source and target domain are identical\cite{Zhang_2017_CVPR}, and 2) certain portion of the data in the source domain can be reused\cite{pan2010survey} for learning in the target domain through re-weighting.
Feature-based adaptation relaxes such a strict assumption and only requires     
 that there exists a mapping from the  input data space to a latent shared feature representation space. This latent shared feature space captures the information necessary for training classifiers for source and target tasks. In this paper, we propose a \textit{feature}-based adaptation DA method. 
 
A common method to approach feature adaptation is to seek a low-dimensional latent subspace\cite{7078994,Busto_2017_ICCV} via dimension reduction.  State of the art  features two main lines of approaches, namely \textit{data geometric structure alignment}-based or \textit{data distribution} centered. Data geometric structure alignment-based approaches, \textit{e.g.}, \textbf{LTSL}\cite{DBLP:journals/ijcv/ShaoKF14} , \textbf{LRSR}\cite{DBLP:journals/tip/XuFWLZ16},  seek a subspace where source and target data can be well aligned and interlaced in preserving inherent hidden geometric data structure via low rank constraint and/or sparse representation.  Data distribution centered methods aim to search a latent subspace where the discrepancy between the source and target data distributions is minimized, via various distances, \textit{e.g.}, Bregman divergence\cite{4967588} based distance, Geodesic distance\cite{gong2012geodesic} or Maximum Mean Discrepancy (MMD) \cite{gretton2012kernel}. The most popular distance is MMD due to its simplicity and solid theoretical foundations. 


A cornerstone theoretical result in DA \cite{ben2010theory,kifer2004detecting} is achieved by  Ben-David \textit{et al.}, who estimate an error bound of a learned hypothesis $h$ on a  target domain:  
%
%
\vspace{-5pt} 
\begin{equation}\label{eq:bound}
		\resizebox{0.50\hsize}{!}{%
			$\begin{array}{l}
			{e_{\cal T}}(h) \le {e_{\cal S}}(h) + {d_{\cal H}}({{\cal D}_{\cal S}},{{\cal D}_{\cal T}})+ \\
			\;\;\;\;\;\;\;\; \min \left\{ {{{\cal E}_{{{\cal D}_{\cal S}}}}\left[ {\left| {{f_{\cal S}}({\bf{x}}) - {f_{\cal T}}({\bf{x}})} \right|} \right],{{\cal E}_{{{\cal D}_{\cal T}}}}\left[ {\left| {{f_{\cal S}}({\bf{x}}) - {f_{\cal T}}({\bf{x}})} \right|} \right]} \right\}
			\end{array}$}
	\end{equation}
	\vspace{-10pt} 
	
	Eq.(\ref{eq:bound})  provides insight on the way to improve DA algorithms as it states that the performance of a hypothesis 
	$h$ on a target domain is determined by: 1) the classification error on the source domain ${e_{\cal S}}(h)$; 2) data divergence ${{d_{\cal H}}({{\cal D}_{\cal S}},{{\cal D}_{\cal T}})}$ which measures the $\mathcal{H}$\emph{-divergence}\cite{kifer2004detecting} between two distributions($\mathcal{D_S}$, $\mathcal{D_T}$); 3) the difference in labeling functions across the two domains. In light of this theoretical result, we can see that  data distribution centered DA methods only seek to minimize the second term in reducing data distribution discrepancies,  whereas data geometric structure alignment-based methods account for the underlying data geometric structure and expect but without theoretical guarantee the alignment of data distributions.  
    
\begin{figure*}[h!]
	\centering
	\includegraphics[width=0.98\linewidth]{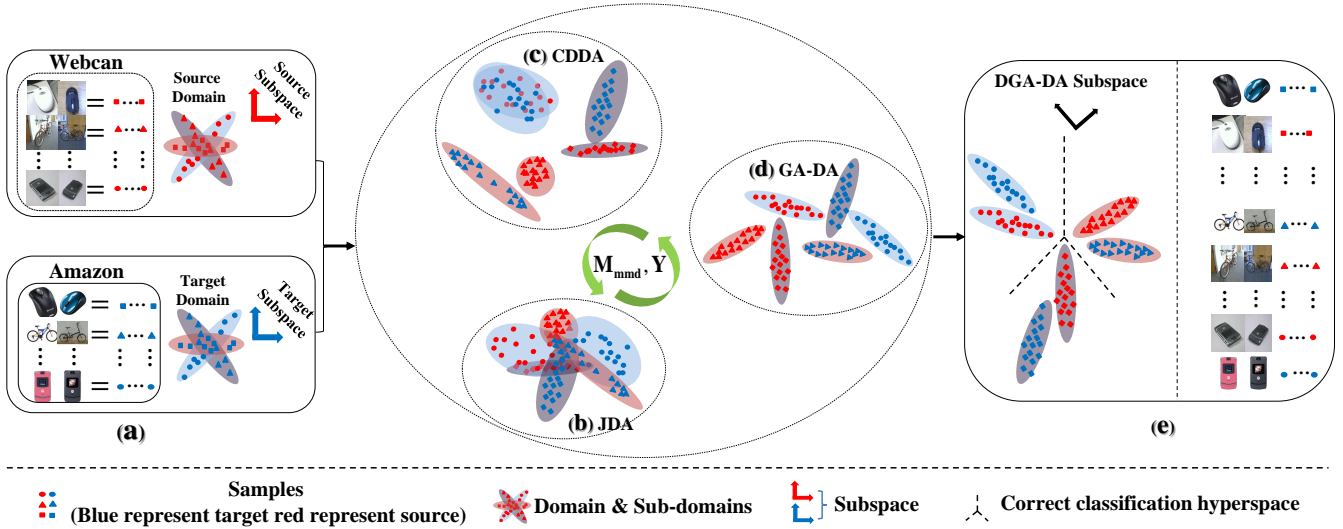}

	\caption {Illustration of the proposed \textbf{DGA-DA} method. Fig.\ref{fig:diff} (a): source data and target data, \textit{e.g.},   mouse, bike, smartphone images, with different distributions and inherent hidden data geometric structures between the source in red  and the target in blue. Samples of different class labels are represented by different geometrical shapes, \textit{e.g.}, round, triangle and square; Fig.\ref{fig:diff} (b) illustrates \textbf{JDA} which closers data distributions whereas \textbf{CDDA} (Fig.\ref{fig:diff} (c)) further makes data discriminative using inter-class repulsive force. Both of them makes use of the nonparametric distance, \textit{i.e.}, Maximum Mean Discrepancy (MMD).	Fig.\ref{fig:diff} (d): accounts for geometric structures of the underlying data manifolds and initial label knowledge in the source domain for label inference; In the proposed DA methods, MMD matrix ${{\bf{M}}_{mmd}}$ and label matrix ${\bf{Y}}$ are updated iteratively within the processes in Fig.\ref{fig:diff} (b-d);  Fig.\ref{fig:diff} (e): the achieved latent joint subspace where both marginal and class conditional data distributions are aligned between source and target as well as their data geometric structures; Furthermore, data instances of different classes are well separated from each other, thereby enabling discriminative DA.} 
	\label{fig:diff}
\end{figure*}

	In this paper, we argue that an effective DA method should:   P1) search a shared feature subspace where source and target data are not only aligned in terms of distributions as most state of the art DA methods do, \textit{e.g.}, \textbf{TCA}\cite{pan2011domain}, \textbf{JDA}\cite{long2013transfer}, but also \textit{discriminative} in that instances of different classes are well separated; P2) account for the  geometric structure of the underlying data manifold when inferring data labels on the target domain.
    
    As a result, we propose in this paper a novel Discriminative Geometry Aware DA (\textbf{DGA-DA}) method which provides  a unified framework for a simultaneous optimization of the three terms in the  upper error bound  in Eq.(\ref{eq:bound}).  Specifically, the proposed \textbf{DGA-DA} also seeks a latent feature subspace to align data distributions as most state of the art DA methods do, but also introduces a \textit{repulsive force} term in the proposed model so as to increase inter-class distances and thereby facilitate discriminative learning and minimize the classification error of the learned hypothesis on source data. Furthermore, the proposed \textbf{DGA-DA} also introduces in its model two additional constraints, namely \textit{Label Smoothness Consistency} and \textit{Geometric Structure Consistency}, to account for the geometric structure of the underlying data manifold when inferring data labels in the target domain, thereby minimizing the third term of the  error bound of the underlying learned hypothesis on the target domain. Fig.\ref{fig:diff} illustrates the proposed DA method.    
    
	
    To gain insight into the proposed method and highlight the contribution of P1) and P2) in comparison with a baseline DA method, \textit{i.e.}, \textbf{JDA} \cite{long2013transfer},  which only cares about data distribution alignment, we further derive two partial DA methods from our DA model, namely Close yet Discriminative DA (\textbf{CDDA}) which implements P1), Geometry Aware DA (\textbf{GA-DA}) based on P2), in addition to our Discriminative and Geometry Aware DA (\textbf{DGA-DA}) which integrates jointly P1) and P2). 
    Comprehensive experiments carried out on standard DA benchmarks, \textit{i.e.},  36 cross-domain image classification tasks through 6 datasets, verify the effectiveness of the proposed method, which consistently outperforms the state-of-the-art DA methods. In-depth analysis using both synthetic data and two 	additional partial models further provide insight into the proposed DA model and highlight its interesting properties. 

To sum up, the contributions of this paper are fourfold: 

\begin{itemize}
	\item We propose a novel \textit{repulsive force} term in the DA model  to increase the discriminative power of the shared latent subspace, aside from narrowing discrepancies of both the marginal and conditional distributions between the source and target domains. 
    
    	\item We introduce \textit{data geometry awareness}, through Label Smoothness  and Geometric Structure Consistencies, for label inference  in the proposed DA model and thereby account for the geometric structures of the underlying data manifold.   
	

	\item We derive from our DA model three novel DA methods, namely \textbf{CDDA}, \textbf{GA-DA} and \textbf{DGA-DA}, which successively implement data discriminativeness, geometry awareness  and both, and quantify the contribution of each term beyond a baseline DA method,\textit{ i.e.}, \textbf{JDA}, which only cares alignment of data distributions. 
	
	\item We perform extensive experiments on 36 image classification DA tasks through 6 popular DA benchmarks and verify the effectiveness of the proposed method which consistently outperforms twenty-two state-of-the-art DA algorithms with a significant margin. Moreover, we also carry out in-depth analysis of the proposed DA methods, in particular \textit{w.r.t.} their hyper-parameters and convergence speed. In addition, using both synthetic and real data, we also provide insights into the proposed DA model in visualizing the effect of data discriminativeness and geometry awareness.

\end{itemize}




The paper is organized as follows. Section 2 discusses the related work. Section 3 presents the method. Section 4 benchmarks the proposed DA method and provides in-depth analysis. Section 5 draws conclusion.   


	\vspace{-5pt} 
\section{Related Work}
Unsupervised Domain Adaptation assumes no labeled data are provided in the target domain. Thus in order to achieve satisfactory classification performance on the target domain, one needs to learn a classifier with labeled samples provided only from the source domain as well as unlabelled samples from the target domain. In earlier days, this problem is also known as \textit{co-variant shift} and can be solved by  sample re-weighting \cite{sugiyama2008direct}. These methods aim to reduce the distribution difference by re-weighting the source samples according to their relevance to the target samples. While proving useful when the data divergence between the source and target domain is small, these methods fall short to align source and target data when this divergence becomes large. 

As a result, recent research in DA has focused its attention on \textit{feature}-based adaptation approach \cite{long2013transfer,Zhang_2017_CVPR,DBLP:journals/ijcv/ShaoKF14,DBLP:conf/cvpr/LongWDSY14,DBLP:journals/tip/XuFWLZ16,DBLP:journals/corr/LuoWHWTC17}, which only assumes a shared latent feature space between the source and target domain. In the learned latent space, the divergence between the projected source and target data distributions is supposed to be minimized. Therefore a classifier learned with the projected labeled source samples could be applied for classification on target samples. To find such a latent shared feature space, many existing methods, \textit{e.g.},\cite{pan2011domain,long2013transfer,Zhang_2017_CVPR,DBLP:conf/cvpr/LongWDSY14,JMLR:v17:15-207}, embrace the dimensionality reduction and propose to explicitly minimize some predefined distance measures to reduce the mismatch between source and target in terms of marginal distribution \cite{4967588} \cite{pan2008transfer} \cite{pan2011domain}, or conditional distribution \cite{satpal2007domain}, or both \cite{long2013transfer}. For example, \cite{4967588} proposed a Bregman Divergence based regularization schema, which combines Bregman divergence with conventional dimensionality reduction algorithms. In \cite{pan2011domain}, the authors use a similar dimensionality reduction framework while making use of the \textit{Maximum Mean Discrepancy} (MMD) based on the Reproducing Hilbert Space (RKHS) \cite{borgwardt2006integrating} to estimate the distance between distributions. In \cite{long2013transfer}, the authors further improve this work by minimizing not only the mismatch of the cross-domain marginal probability distributions, but also  the mismatch of conditional probability distributions. 

In line with the focus of manifold learning \cite{Zhou04learningwith}, an increasing number of DA methods, \textit{e.g.}, \cite{DBLP:journals/corr/LuoWHC17,DBLP:journals/ijcv/ShaoKF14,DBLP:journals/tip/XuFWLZ16},  emphasize the importance of aligning the underlying data manifold structures between the source and the target domain for effective DA. In these methods, low-rank and sparse constraints are introduced into DA to extract a low-dimension feature subspace where target samples can be sparsely reconstructed from source samples \cite{DBLP:journals/ijcv/ShaoKF14}, or interleaved by source samples \cite{DBLP:journals/tip/XuFWLZ16},  thereby aligning the geometric structures of the underlying data manifolds. A few recent DA methods, \textit{e.g.}, \textbf{RSA-CDDA}\cite{DBLP:journals/corr/LuoWHC17}, \textbf{JGSA}\cite{Zhang_2017_CVPR}, further propose unified frameworks to reduce the shift between domains both statistically and geometrically.  

However, in light of the upper error bound as defined in Eq.(\ref{eq:bound}), we can see that data distribution centered DA methods only seek to minimize the second term in reducing data distribution discrepancies,  whereas data geometric structure alignment-based methods account for the underlying data geometric structure and expect but without theoretical guarantee the alignment of data distributions. In contrast, the proposed \textbf{DGA-DA} method optimizes altogether the three error terms of the upper error bound in Eq.(\ref{eq:bound}). 

The proposed \textbf{DGA-DA} builds on \textbf{JDA} \cite{long2013transfer} in seeking a latent feature subspace while minimizing the mismatch of both the marginal and conditional probability distributions across domains, thereby decreasing the data divergence term in Eq.(\ref{eq:bound}). But \textbf{DGA-DA} goes beyond and differs from \textbf{JDA} as we introduce in the proposed DA model a \textit{repulsive force} term so as to increase inter-class distances for discriminative DA, thereby optimizing the first term of the upper error bound in Eq.(\ref{eq:bound}), \textit{i.e.},  the error rate of the learned hypothesis on the source domain. Furthermore, the proposed \textbf{DGA-DA} also accounts in its model for the  geometric structures of the underlying data manifolds, through label smoothness consistency (LSC) and geometric structure consistency (GSC) which require the inferred labels on the source and target data be smooth and have similar labels on nearby data. These two constraints thus further optimize the third term of the upper error bound in Eq.(\ref{eq:bound}). \textbf{DGA-DA} also differs much from a recent DA method, \textit{i.e.}, \textbf{SCA}\cite{DBLP:journals/pami/GhifaryBKZ17}, which also tries to introduce data discriminativeness through the between and within class scatter only defined on the source domain. However, besides data geometry awareness that it does not consider, \textbf{SCA} does not seek explicitly data distribution alignment as we do in heritage of \textbf{JDA}, nor it has the \textit{repulsive force} term as we introduce in our model in pushing away inter-class data based on both source and target domain.  Using both synthetic and real data, sect.\ref{subsection:Analysis and Verification} provides insights into and visualizes the differences of the proposed model with a number of state of the art DA methods, \textit{e.g.}, \textbf{SCA},  and highlights its interesting properties, in particular data distribution alignment, data discriminativeness and geometry awareness.

\section{Discriminative   Geometry Aware Domain Adaptation}
We first introduce the notations and formalize the problem in sect.\ref{subsection:Notations and Problem Statement}, then present in sect.\ref{subsection:the model}  the proposed model for Discriminative and Geometry Aware Domain Adaptation (\textbf{DGA-DA}), and solve the model in sect.\ref{subsection:solving the model}. Sect.\ref{ssection:Kernelization Analysis} further analyzes the kernelization of the proposed DA model for nonlinear DA problems.

\subsection{Notations and Problem Statement}
\label{subsection:Notations and Problem Statement}

Matrices are written as boldface uppercase letters. Vectors are written as boldface lowercase letters. For matrix ${\bf{X}} = ({x_{ij}})$, its $i$-th row is denoted as ${{\bf{x}}^i}$, and its $j$-th column is denoted by ${{\bf{x}}_j}$.  We define the Frobenius norm ${\left\| . \right\|_F}$ as: ${\left\| {\bf{X}} \right\|_F} = \sqrt {\sum {_{i = 1}^n} \sum {_{j = 1}^m} x_{ij}^2} $ . 
	
	A domain $D$ is defined as an m-dimensional feature space $\chi$ and a marginal probability distribution $P(x)$, \textit{i.e.}, $\mathcal{D}=\{\chi,P(x)\}$ with $x\in \chi$.  Given a specific domain $D$, a  task $T$ is composed of a C-cardinality label set $\mathcal{Y}$  and a classifier $f(x)$,\textit{ i.e.}, $T = \{\mathcal{Y},f(x)\}$, where $f({x}) = \mathcal{Q}( y |x)$ can be interpreted as the class conditional probability distribution for each input sample $x$.

	In unsupervised domain adaptation, we are given a source domain $\mathcal{D_S}=\{x_{i}^{s},y_{i}^{s}\}_{i=1}^{n_s}$ with $n_s$ labeled samples ${{\bf{X}}_{\cal S}} = [x_1^s...x_{{n_s}}^s]$, which are associated with their class labels ${{\bf{Y}}_S} = {\{ {y_1},...,{y_{{n_s}}}\} ^T} \in {{\bf{\mathbb{R}}}^{{n_s} \times c}}$, and an unlabeled target domain $\mathcal{D_T}=\{x_{j}^{t}\}_{j=1}^{n_t}$ with $n_t$  unlabeled samples ${{\bf{X}}_{\cal T}} = [x_1^t...x_{{n_t}}^t]$, whose labels are ${{\bf{Y}}_T} = {\{ {y_{{n_s} + 1}},...,{y_{{n_s} + {n_t}}}\} ^T} \in {{\bf{\mathbb{R}}}^{{n_t} \times c}}$ are unknown. Here, source domain labels  ${y_i} \in {{\bf{\mathbb{R}}}^c}(1 \le i \le {n_s})$ is a binary vector in which $y_i^j = 1$ if ${x_i}$ belongs to the $j$-th class. We  define the data matrix ${\bf{X}} = [{{\bf{X}}_S},{{\bf{X}}_T}] \in {R^{m*n}}$ in packing both the source and target data. The source domain $\mathcal{D_S}$ and target domain $\mathcal{D_T}$ are assumed to be different, \textit{i.e.},  $\mathcal{\chi}_S=\mathcal{{\chi}_T}$, $\mathcal{Y_S}=\mathcal{Y_T}$, $\mathcal{P}(\mathcal{\chi_S}) \neq \mathcal{P}(\mathcal{\chi_T})$, $\mathcal{Q}(\mathcal{Y_S}|\mathcal{\chi_{S}}) \neq \mathcal{Q}(\mathcal{Y_T}|\mathcal{\chi_{T}})$.

	We also define the notion of \textit{sub-domain}, denoted as ${\cal D}_{\cal S}^{(c)}$, representing the set of samples in ${{\cal D}_{\cal S}}$ with the label $c$. Similarly, a sub-domain ${\cal D}_{\cal T}^{(c)}$ can be defined for the target domain as the set of samples in ${{\cal D}_{\cal T}}$ with the label $c$. However, as samples in the target domain ${{\cal D}_{\cal T}}$ are unlabeled, the definition of sub-domains in the target domain, requires a base classifier,\textit{ e.g.}, Nearest Neighbor (NN),  to attribute  pseudo labels for samples in ${{\cal D}_{\cal T}}$.

	The  maximum mean discrepancy (MMD)  is an effective non-parametric distance-measure  that compares the distributions of two sets of data by mapping the data to Reproducing Kernel Hilbert Space\cite{borgwardt2006integrating} (RKHS). Given two distributions $\mathcal{P}$ and $\mathcal{Q}$, the MMD between $\mathcal{P}$ and $\mathcal{Q}$ is defined as:
	\begin{equation}
		\label{eq:MMD}
		Dist(P,Q) = \parallel \frac{1}{n_1} \sum^{n_1}_{i=1} \phi(p_i) - \frac{1}{n_2} \sum^{n_2}_{i=1} \phi(q_i) \parallel_{\mathcal{H}}
	\end{equation}
	where $P=\{ p_1, \ldots, p_{n_1} \}$ and $Q = \{ q_1, \ldots, q_{n_2} \}$ are two random variable sets from distributions $\mathcal{P}$ and $\mathcal{Q}$, respectively, and $\mathcal{H}$ is a universal RKHS with the reproducing kernel mapping $\phi$: $f(x) = \langle \phi(x), f \rangle$, $\phi: \mathcal{X} \to \mathcal{H}$.

The aim of the Discriminative and Geometry Aware Domain Adaptation (\textbf{DGA-DA}) is to learn a latent feature subspace with the following properties: P1) the distances of both marginal and conditional probabilities between the source and target domains are reduced; P2)  The distances between each sub-domain to the others are increased so as to  increase inter-class distances and thereby enable discriminative DA; and P3) label inference accounts for the underlying data geometric structure.

\subsection{The model}

\label{subsection:the model}
The proposed DA model (sect.\ref{subsubsection: the final model (DGA-DA)}) builds on \textbf{TCA} (sect.\ref{subsubsection: TCA}) and \textbf{JDA} (sect.\ref{subsubsection:JDA}) to which discriminative DA (\textbf{CDDA}) is introduced (sect.\ref{subsubsection:Discriminative DA}) and the data geometry awareness (\textbf{GA-DA}) is accounted for in label inference and the search of the shared latent feature subspace (sect.\ref{subsubsection:GA-DA}).

\subsubsection{Search of a Latent Feature Space with Dimensionality Reduction (\textbf{TCA})}
\label{subsubsection: TCA}
The search of a latent feature subspace  with dimensionality reduction  has been demonstrated useful for DA in several previous works, \textit{e.g.}, \cite{pan2011domain,long2013transfer,DBLP:journals/corr/LuoWHC17,DBLP:journals/ijcv/ShaoKF14,Zhang_2017_CVPR}.  In projecting  original raw data into a lower dimensional space,  the \emph{principal} data structure is preserved while decreasing its complexities. In the proposed method, we also apply the Principal Component Analysis (PCA) to capture the major data structure.  Mathematically, given  an input data matrix $\boldsymbol{X} = [{\mathcal{D_S}},\mathcal{D_T}]$, $\boldsymbol{X} \in {\mathbb{R}^{m\times({n_s} + {n_t})}}$, the centering matrix is defined as  $\boldsymbol{H} = \boldsymbol{I} - \frac{1}{n_s+n_t}\boldsymbol{1}$, where $\boldsymbol{1}$ is the $(n_s+n_t) \times (n_s+n_t)$ matrix of ones. The optimization of PCA is to find a projection transformation $\boldsymbol{A}$ which  maximizes the embedded data variance.
\begin{equation}\label{eq:pca}
	\begin{array}{c}
		\mathop {\max}\limits_{\boldsymbol{A^TA} = \boldsymbol{I}} tr(\boldsymbol{A}^T\boldsymbol{ XH}\boldsymbol{X}^T \boldsymbol{A})
	\end{array}
\end{equation}
where $tr(\mathord{\cdot})$ denotes the trace of a matrix,   $\boldsymbol{XH}\boldsymbol{X}^T$ is the data covariance matrix, and $\bf A \in \mathbb{R}^{m \times k}$ with $m$ the feature dimension and $k$ the dimension of the projected subspace. The optimal solution  is calculated by solving an eigendecomposition problem: $\boldsymbol{XH}\boldsymbol{X}^T=\boldsymbol{A\Phi}$, where $\boldsymbol{\Phi}=diag(\phi_1,\dots, \phi_k )$ are the $k$ largest eigenvalues. Finally, the original data $\boldsymbol{X}$ is projected into the  optimal $k$-dimensional subspace using $\boldsymbol{Z} = \boldsymbol{A}^T\boldsymbol{X}$.

\subsubsection{Joint Marginal and Conditional Distribution Domain Adaptation (\textbf{JDA})}
\label{subsubsection:JDA}
However, the previous feature subspace calculated via PCA does not align explicitly data distributions between the source and target domain.  Following \cite{long2013transfer,long2015learning}, we also empirically measure the distance of both marginal and conditional distributions across domain  via the nonparametric distance measurement MMD in RKHS \cite{borgwardt2006integrating} once the original data projected into  a low-dimensional feature space.  Formally, the empirical distance of the two domains  is defined as:
\begin{equation}\label{eq:marginal}
		\resizebox{0.50\hsize}{!}{%
	$\begin{array}{l}		
		Dis{t^{marginal}}({{\cal D}_{\cal S}},{{\cal D}_{\cal T}}) =\\ {\left\| {\frac{1}{{{n_s}}}\sum\limits_{i = 1}^{{n_s}} {{{\bf{A}}^T}{x_i} - } \frac{1}{{{n_t}}}\sum\limits_{j = {n_s} + 1}^{{n_s} + {n_t}} {{{\bf{A}}^T}{x_j}} } \right\|^2}
		= tr({{\bf{A}}^T}\bf{X}{\bf{M_0}}\bf{{X^T}A})		
	\end{array}$}
\end{equation}

where ${{\bf{M}}_0}$ represents the marginal distribution between ${{\cal D}_{\cal S}}$ and ${{\cal D}_{\cal T}}$ and its calculation is obtained by:

\begin{equation}\label{eq:M0}
		\resizebox{0.4\hsize}{!}{%
$\begin{array}{*{20}{l}}
{{{({{\bf{M}}_0})}_{ij}} = \left\{ {\begin{array}{*{20}{l}}
		{\frac{1}{{{n_s}{n_s}}},\;\;\;{x_i},{x_j} \in {D_S}}\\
		{\frac{1}{{{n_t}{n_t}}},\;\;\;{x_i},{x_j} \in {D_T}}\\
		{\frac{{ - 1}}{{{n_t}{n_s}}},\;\;\;\;\;\;\;\;\;\;\;\;otherwise}
		\end{array}} \right.}
\end{array}$}
\end{equation}


where ${x_i},{x_j} \in (\mathcal{D_S} \cup \mathcal{D_T})$. The difference between the marginal distributions $\mathcal{P}(\mathcal{X_S})$ and $\mathcal{P}(\mathcal{X_T})$ is reduced in minimizing {$Dis{t^{marginal}}({{\cal D}_{\cal S}},{{\cal D}_{\cal T}})$}.

Similarly, the distance of conditional probability distributions is defined as the sum of the empirical distances over the class labels between the sub-domains of a same label in the source and target domain: 

\begin{equation}\label{eq:conditional}
		\resizebox{1\hsize}{!}{%
$\begin{array}{c}
		\begin{array}{l}
			Dis{t^{conditional}}\sum\limits_{c = 1}^C {({{\cal D}_{\cal S}}^c,{{\cal D}_{\cal T}}^c)}  = 			{\left\| {\frac{1}{{n_s^{(c)}}}\sum\limits_{{x_i} \in {{\cal D}_{\cal S}}^{(c)}} {{{\bf{A}}^T}{x_i}}  - \frac{1}{{n_t^{(c)}}}\sum\limits_{{x_j} \in {{\cal D}_{\cal T}}^{(c)}} {{{\bf{A}}^T}{x_j}} } \right\|^2}= tr({{\bf{A}}^T}{\bf{X}}{{\bf{M}}_c}{{\bf{X}}^{\bf{T}}}{\bf{A}})
		\end{array}
	\end{array}$}
\end{equation}
where $C$ is the number of classes, $\mathcal{D_S}^{(c)} = \{ {x_i}:{x_i} \in \mathcal{D_S} \wedge y({x_i}) = c\} $ represents the ${c^{th}}$ sub-domain in the source domain, $n_s^{(c)} = {\left\| {\mathcal{D_S}^{(c)}} \right\|_0}$ is the number of samples in the ${c^{th}}$ {source} sub-domain. $\mathcal{D_T}^{(c)}$ and $n_t^{(c)}$ are defined similarly for the target domain. Finally, $\bf M_c$ represents the conditional distribution between sub-domains in ${{\cal D}_{\cal S}}$ and ${{\cal D}_{\cal T}}$ and it is defined as: 
\begin{equation}\label{eq:Mc}
		\resizebox{0.45\hsize}{!}{%
	$\begin{array}{*{20}{c}}
		{{{({{\bf{M}}_c})}_{ij}} = \left\{ {\begin{array}{*{20}{l}}
					{\frac{1}{{n_s^{(c)}n_s^{(c)}}},\;\;\;{x_i},{x_j} \in {D_{\cal S}}^{(c)}}\\
					{\frac{1}{{n_t^{(c)}n_t^{(c)}}},\;\;\;{x_i},{x_j} \in {D_{\cal T}}^{(c)}}\\
					{\frac{{ - 1}}{{n_s^{(c)}n_t^{(c)}}},\;\;\;\left\{ {\begin{array}{*{20}{l}}
								{{x_i} \in {D_{\cal S}}^{(c)},{x_j} \in {D_{\cal T}}^{(c)}}\\
								{{x_i} \in {D_{\cal T}}^{(c)},{x_j} \in {D_{\cal S}}^{(c)}}
							\end{array}} \right.}\\
						{0,\;\;\;\;\;\;\;\;\;\;\;\;otherwise}
					\end{array}} \right.}
			\end{array}$}
		\end{equation}
		
		
In minimizing ${Dis{t^{conditional}}\sum\limits_{c = 1}^C {({D_{\cal S}}^c,{D_{\cal T}}^c)} }$,  the mismatch of conditional distributions between ${{D_{\cal S}}^c}$ and ${{D_{\cal T}}^c}$ is reduced.

\subsubsection{Close yet Discriminative Domain Adaptation (\textbf{CDDA})}\label{subsubsection:Discriminative DA}
	
However, the previous joint alignment of the marginal and conditional distributions across domain does not explicitly render data discriminative  in the searched feature subspace. As a result, we introduce a  \textit{Discriminative} domain adaption via a \textit{repulsive force} term, so as to increase the distances of sub-domains with different labels, and improve the discriminative power of the latent shared features, thereby making it possible for a better predictive model for both the source and target data. 
    
		
Specifically, the \textit{repulsive force} term is defined as: 
$Dis{t^{repulsive}} = Dist_{{\cal S} \to {\cal T}}^{repulsive} + Dist_{{\cal T} \to {\cal S}}^{repulsive}$, where ${{\cal S} \to {\cal T}}$ and ${{\cal T} \to {\cal S}}$ index the distances computed from ${D_{\cal S}}$ to ${D_{\cal T}}$ and ${D_{\cal T}}$ to ${D_{\cal S}}$, respectively. $Dist_{{\cal S} \to {\cal T}}^{repulsive}$ represents the sum of the distances between each source sub-domain ${D_{\cal S}}^{(c)}$ and all the  target sub-domains ${D_{\cal T}}^{(r);\;r \in \{ \{ 1...C\}  - \{ c\} \} }$ except the one with the label $c$. The sum of these distances is explicitly defined as:
		\begin{equation}\label{eq:StoT}
			\resizebox{1\hsize}{!}{%
				${Dist}_{{\cal S} \to {\cal T}}^{repulsive} = \sum\limits_{c = 1}^C \begin{array}{l}
				{\left\| {\frac{1}{{n_s^{(c)}}}\sum\limits_{{x_i} \in {D_{\cal S}}^{(c)}} {{{\bf{A}}^T}{x_i}}  - \frac{1}{{\sum\limits_{r \in \{ \{ 1...C\}  - \{ c\} \} } {n_t^{(r)}} }}\sum\limits_{{x_j} \in D_{\cal T}^{(r)}} {{{\bf{A}}^T}{x_j}} } \right\|^2}
				= \sum\limits_{c = 1}^C {tr({{\bf{A}}^T}{\bf{X}}{{\bf{M}}_{{\cal S} \to {\cal T}}}{{\bf{X}}^{\bf{T}}}{\bf{A}})} 
				\end{array} $}
		\end{equation}
where ${{\bf{M}}_{{\cal S} \to {\cal T}}}$ is defined as
\begin{equation}\label{eq:mstot}
		\resizebox{0.65\hsize}{!}{%
			$\begin{array}{c}
				(\bf M_{{{\cal S} \to {\cal T}}})_{ij} = \left\{ {\begin{array}{*{20}{l}}
						{\frac{1}{{n_s^{(c)}n_s^{(c)}}},\;\;\;{x_i},{x_j} \in {D_{\cal S}}^{(c)}}\\
						{\frac{1}{{n_t^{(r)}n_t^{(r)}}},\;\;\;{x_i},{x_j} \in {D_{\cal T}}^{(r)}}\\
						{\frac{{ - 1}}{{n_s^{(c)}n_t^{(r)}}},\;\;\;\left\{ {\begin{array}{*{20}{l}}
									{{x_i} \in {\cal D_{\cal S}}^{(c)},{x_j} \in {D_{\cal T}}^{(r)}}\\
									{{x_i} \in {\cal D_{\cal T}}^{(r)},{x_j} \in {\cal D_{\cal S}}^{(c)}}
								\end{array}} \right.}\\
							{0,\;\;\;\;\;\;\;\;\;\;\;\;otherwise}
						\end{array}} \right.
					\end{array}$}
				\end{equation}
Symmetrically, $Dist_{{\cal T} \to {\cal S}}^{repulsive}$ represents the sum of the distances from each target sub-domain ${D_{\cal T}}^{(c)}$ to all the the source sub-domains ${D_{\cal S}}^{(r);\;r \in \{ \{ 1...C\}  - \{ c\} \} }$ except the source sub-domain with the label $c$. Similarly, the sum of these distances is explicitly defined as:	
\begin{equation}\label{eq:TtoS}
					\resizebox{1\hsize}{!}{%
						$Dist_{T \to S}^{repulsive} = \sum\limits_{c = 1}^C \begin{array}{l}
						{\left\| {\frac{1}{{n_s^{(c)}}}\sum\limits_{{x_i} \in {D_T}^{(c)}} {{{\bf{A}}^T}{x_i}}  - \frac{1}{{\sum\limits_{r \in \{ \{ 1...C\}  - \{ c\} \} } {n_t^{(r)}} }}\sum\limits_{{x_j} \in D_S^{(r)}} {{{\bf{A}}^T}{x_j}} } \right\|^2}
						= \sum\limits_{c = 1}^C {tr({{\bf{A}}^T}{\bf{X}}{{\bf{M}}_{T \to S}}{{\bf{X}}^{\bf{T}}}{\bf{A}})} 
						\end{array}  $}
				\end{equation}
where ${{\bf{M}}_{{\cal T} \to {\cal S}}}$ is defined as
				\begin{equation}\label{eq:mttos}
                		\resizebox{0.65\hsize}{!}{%
					$\begin{array}{c}
						(\bf M_{{{\cal T} \to {\cal S}}})_{ij} = \left\{ {\begin{array}{*{20}{l}}
								{\frac{1}{{n_t^{(c)}n_t^{(c)}}},\;\;\;{x_i},{x_j} \in {D_{\cal T}}^{(c)}}\\
								{\frac{1}{{n_s^{(r)}n_s^{(r)}}},\;\;\;{x_i},{x_j} \in {D_{\cal S}}^{(r)}}\\
								{\frac{{ - 1}}{{n_t^{(c)}n_s^{(r)}}},\;\;\;\left\{ {\begin{array}{*{20}{l}}
											{{x_i} \in {\cal D_{\cal T}}^{(c)},{x_j} \in {D_{\cal S}}^{(r)}}\\
											{{x_i} \in {\cal D_{\cal S}}^{(r)},{x_j} \in {\cal D_{\cal T}}^{(c)}}
										\end{array}} \right.}\\
									{0,\;\;\;\;\;\;\;\;\;\;\;\;otherwise}
								\end{array}} \right.
							\end{array}$}
						\end{equation}
						
Finally, we obtain
		\begin{equation}\label{eq:repulsive}
							\resizebox{0.55\hsize}{!}{%
								${Dist}^{repulsive} = \sum\limits_{c = 1}^C {tr({{\bf{A}}^T}{\bf{X}}({{\bf{M}}_{S \to T}} + {{\bf{M}}_{T \to S}}){{\bf{X}}^{\bf{T}}}{\bf{A}})} $}
						\end{equation}
						
We define ${{\bf{M}}_{\hat c}} = {{\bf{M}}_{S \to T}} + {{\bf{M}}_{T \to S}}$ as the \textit{repulsive force} matrix.
While the minimization of Eq.(\ref{eq:marginal}) and Eq.(\ref{eq:conditional}) makes closer both marginal and conditional distributions between source and target, the maximization of Eq.(\ref{eq:repulsive}) increases the  distances between source and target sub-domains, thereby improve the discriminative power of the searched latent feature subspace. 
						

\subsubsection{Geometry Aware Domain Adaptation (\textbf{GA-DA})}
\label{subsubsection:GA-DA}
In a number of state of the art DA methods, \textit{e.g.},\cite{pan2008transfer,pan2011domain,long2013transfer}, the simple \textit{Nearest Neighbor} (NN) classifier is applied for label inference. In \textbf{JDA} and \textbf{LRSR}\cite{DBLP:journals/tip/XuFWLZ16}, NN-based label deduction is applied twice at each iteration. NN is first applied to the target domain in order to generate the \emph{pseudo} labels of the target data and enable the computation of the conditional probability distance as defined in sect. \ref{subsubsection:JDA}. Once the optimized latent subspace  identified, NN is then applied once again at the end of an iteration for the label prediction of the target domain. However, given the neighborhood  usually based on the $L2$ or $L1$ distance, NN could fall short to measure the similarity of source and target domain data which may be embedded into a manifold with complex geometric structures.


To account for the underlying data manifold structure in data similarity measurement, we further introduce two consistency constraints, namely \textit{label smoothness consistency} and \textit{geometric structure consistency} for both the \emph{pseudo} and final label inference. 
						
						

\textbf{Label Smoothness Consistency (LSC)}：LSC is a constraint designed to prevent too much changes from the initial query assignment ${{\bf{Y}}_{\cal S}}$.

\begin{equation}\label{eq:labelconsitency}
	\begin{array}{c}
Dis{t^{lable}} = \sum\limits_{j = 1}^C {\mathop \sum \limits_{i = 1}^{{n_s} + {n_t}} } \left\| {\bf{Y}}^{(F)}_{i,j}-{\bf{Y}}^{(0)}_{i,j} \right\|
	\end{array}
\end{equation}

where ${\bf{Y}} = {{\bf{Y}}_{\cal S}} \cup {{\bf{Y}}_{\cal T}}$,  ${\bf{Y}}_{i,j}^{(F)}$ is the calculated probability of ${i_{th}}$ data belonging to ${j_{th}}$ class. Each data ${x_i}$ has a predicted label ${y_i} = \arg {\max _{j \le c}}{\bf{Y}}_{ij}^F$. ${\bf{Y}}_{i,j}^{(0)}$ is the initial prediction. As for unlabeled target data ${{\bf{X}}_{\cal T}}$,  traditional ranking methods\cite{6341755,6619251} assign the labels ${{\bf{Y}}_T} = {{\bf{0}}^{{n_t}*c}}$. However, this definition lacks  discriminative properties due to the equal probability assignments in ${{\bf{X}}_{\cal T}}$. In this work, we define the initial ${{\bf{Y}}^{(0)}}$ as:

\begin{equation}\label{eq:labelconsistency1}
		\resizebox{0.55\hsize}{!}{%
	$\begin{array}{*{20}{l}}
	{{\bf{Y}}_{{{\cal S}_{(ij)}}}^{(0)} = \left\{ {\begin{array}{*{20}{l}}
	{y_{{{\cal S}_{(ij)}}}^{(0)} = 1\;(1 \le i \le {n_s}),j = c,{y_{ij}} \in D_{\cal S}^{(c)}}\\
	{0\;\;\;\;\;\;\;\;\;\;\;else}
 \end{array}} \right.}\\
	{{\bf{Y}}_{{{\cal T}_{(ij)}}}^{(0)} = \left\{ {\begin{array}{*{20}{l}}
	\begin{array}{l}
	y_{{{\cal T}_{(ij)}}}^{(0)} = 1\;(({n_s} + 1) \le i \le {n_s} + {n_t}),j = c,\\
	{y_{ij}} \in D_{\cal T}^{(c)}
	\end{array}\\
{0\;\;\;\;\;\;\;\;\;\;\;else}
	\end{array}} \right.}
	\end{array}$}
\end{equation}
	
 where	$D_{\cal T}^{(c)}$ is defined as pseudo labels,  generated via a base classifier, \textit{e.g.}, NN.

\textbf{Geometric Structure Consistency (GSC)}:  GSC is designed to ensure that inferred data labels comply with the  geometric structures of the underlying data manifolds. We propose to characterize alignment of label inference with the underlying data geometric structure through the Laplace matrix $\textbf{L}$:
	\vspace{-1pt}
		\begin{equation}\label{eq:YLY}
        		\resizebox{1\hsize}{!}{%
					$\begin{array}{c}
										\begin{array}{l}
								\begin{array}{l}
{{\bf{Y}}^T}{\bf{L}}{\bf{Y}} = {{\bf{Y}}^T}({\bf{I}} - {{\bf{D}}^{ - \frac{1}{2}}}{\bf{W}}{{\bf{D}}^{ - \frac{1}{2}}}){\bf{Y}} = 
\sum\limits_{i = 1}^{{n_s} + {n_t}} {{d_{ii}}{{\left( {\frac{{{y_i}}}{{\sqrt {{{\bf{d}}_{ii}}} }}} \right)}^2}}  - \sum\limits_{i,j = 1}^{{n_s} + {n_t}} {{{\bf{d}}_{ii}}{{\left( {\frac{{{y_i}}}{{\sqrt {{{\bf{d}}_i}} }}\frac{{{y_j}}}{{\sqrt {{{\bf{d}}_j}} }}} \right)}^2}} {{\bf{w}}_{ij}}\;
= \frac{1}{2}\sum\limits_{i,j = 1}^{{n_s} + {n_t}} {{{\bf{w}}_{ij}}{{\left( {\frac{{{y_i}}}{{\sqrt {{{\bf{d}}_{ii}}} }} - \frac{{{y_j}}}{{\sqrt {{{\bf{d}}_{jj}}} }}} \right)}^2}} 
\end{array}
\end{array},
\end{array}$}
\end{equation}

where  ${\bf{W}} = {[{w_{ij}}]_{({n_s} + {n_t}) \times ({n_s} + {n_t})}}$ is an affinity matrix \cite{NIPS2001_2092}, with   ${w_{ij}}$ giving the affinity between two data samples $i$ and $j$ and  defined as ${w_{ij}} = \exp ( - \frac{{{{\left\| {{x_i} - {x_j}} \right\|}^2}}}{{2{\sigma ^2}}})$ if $i \ne j$ and ${w_{ii}} = 0$ otherwise, ${\bf{D}} = diag\{ {d_{11}}...{d_{({n_s} + {n_t}),({n_s} + {n_t})}}\} $ is the degree matrix with ${d_{ii}} = \sum\nolimits_j {{w_{ij}}} $. When Eq.(\ref{eq:YLY}) is minimized, the geometric structure consistency ensures that the label space does not change too much between nearby data.

\subsubsection{the final model (\textbf{DGA-DA})}
\label{subsubsection: the final model (DGA-DA)}
Our final DA model integrates: 1) alignment of both marginal and conditional distributions across domain as defined by Eq.(\ref{eq:marginal}) and Eq.(\ref{eq:conditional}), 2) the repulsive force as in Eq.(\ref{eq:repulsive}), and 3) data geometry aware label inference through  both the  label smoothness (Eq.(\ref{eq:labelconsitency})) and geometric structure (Eq.(\ref{eq:YLY})) consistencies. Therefore, our final model is defined as:
								\begin{equation}\label{eq:ours_physical}
									\resizebox{0.8\hsize}{!}{%
										$\begin{array}{*{20}{l}}
										{\min (Dis{t^{marginal}} + Dis{t^{conditional}} + Dis{t^{label}} + {{\bf{Y}}^T}L{\bf{Y}})} + 
										\max (Dist^{repulsive})
										\end{array}$}
								\end{equation}
		\vspace{-2pt}					
It can be re-written mathematically as:
	\begin{equation}\label{eq:final model}
	\resizebox{1\hsize}{!}{%
		$\begin{array}{*{20}{l}}
		{\begin{array}{*{20}{l}}
			{\mathop {\min }\limits_{{{\bf{A}}^T}{\bf{XH}}{{\bf{X}}^T}{\bf{A}} = {\bf{I}}} \left( {\begin{array}{*{20}{l}}
					{\sum\limits_{c = 0}^C {tr({{\bf{A}}^T}{\bf{X}}{{\bf{M}}_c}{{\bf{X}}^T}A)}  + \lambda \left\| {\bf{A}} \right\|_F^2}
					{ + \mu (\sum\limits_{j = 1}^C {\sum\limits_{i = 1}^{{n_s} + {n_t}} {\left\| {{\bf{Y}}_{ij}^{(F)} - {\bf{Y}}_{ij}^{(0)}} \right\|} } ) + {{\bf{Y}}^T}{\bf{LY}}}
					\end{array}} \right)}
			{ + \mathop {\max }\limits_{{{\bf{A}}^T}{\bf{XH}}{{\bf{X}}^T}{\bf{A}} = {\bf{I}}} tr({{\bf{A}}^T}{\bf{X}}{{\bf{M}}_{{\bf{\hat c}}}}{{\bf{X}}^T}{\bf{A}})}
			\end{array}}
		\end{array}$}
\end{equation}
\vspace{-2pt}
where the constraint ${{{\bf{A}}^T}{\bf{XH}}{{\bf{X}}^T}{\bf{A}} = {\bf{I}}}$ removes an arbitrary scaling factor in the embedding and prevents the above optimization collapse onto a subspace of dimension less than the required dimensions. $\lambda$ is a regularization parameter to guarantee the optimization problem to be well-defined. $\mu $ is a trade-off parameter which balances LSC and GSC.
\vspace{-3pt}				
\subsection{Solving the model}
\label{subsection:solving the model}

Direct solution to Eq.(\ref{eq:final model}) is nontrivial. We divide it into two sub-problems. 

\textbf{Sub-problem (a)}:
\begin{equation}\label{eq:prob1}
		\resizebox{0.7\hsize}{!}{%
$\mathop {\min }\limits_{{{\bf{A}}^T}{\bf{XH}}{{\bf{X}}^T}{\bf{A}} = {\bf{I}},{{\bf{M}}_{cyd}} = \sum\limits_{c = 0}^C {{{\bf{M}}_c} - {{\bf{M}}_{{\bf{\hat c}}}}} } \left( {\sum\limits_{c = 0}^C {tr({{\bf{A}}^T}{\bf{X}}{{\bf{M}}_{cyd}}{{\bf{X}}^T}A)}  + \lambda \left\| {\bf{A}} \right\|_F^2} \right),$}
\end{equation}

\textbf{Sub-problem (b)}:
\begin{equation}\label{eq:prob2}
		\resizebox{0.5\hsize}{!}{%
$\min \left( {\mu \sum\limits_{j = 1}^C {\sum\limits_{i = 1}^{{n_s} + {n_t}} {\left\| {{\bf{Y}}_{ij}^{(F)} - {\bf{Y}}_{ij}^{(0)}} \right\|} }  + {{\bf{Y}}^T}{\bf{LY}}} \right)$}
\end{equation}
  
These two sub-problems are then iteratively optimized. 
								
Sub-problem (a) amounts to solving the generalized eigendecomposition problem. Augmented Lagrangian method \cite{fortin2000augmented,long2013transfer} can be used to solve this problem. In setting its partial derivation \textit{w.r.t.} $\boldsymbol{A}$ equal to zero, we obtain:

\begin{equation}\label{eq:eig}
({\bf{X}}{\bf{M_{cyd}}}{{\bf{X}}^T} + \lambda {\bf{I}}){\bf{A}} = {\bf{XH}}{{\bf{X}}^T}{\bf{A}}\Phi 
\end{equation}

where $\Phi {\rm{ = diagram}}({\varphi _1},...{\varphi _k}) \in {R^{k*k}}$ is the Lagrange multiplier. The optimal subspace $\boldsymbol{A}$ is reduced to solving Eq.(\ref{eq:eig}) for the k smallest eigenvectors.  Then, we obtain the projection matrix $\boldsymbol{A}$ and the underlying embedding space ${\bf{Z}} = {{\bf{A}}^T}{\bf{X}}$. 

Sub-problem (b) is  nontrivial. Inspired by the solution proposed in   \cite{Zhou04learningwith} \cite{6341755} \cite{6619251}, the minimum is approached where the derivative of the function is zero.  An approximate solution can be provided by: 
\begin{equation}\label{eq:Y_optimal}
{{\bf{Y}}^ \star } = {({\bf{D}} - (\frac{1}{{1 + \mu }}){\bf{W}})^{ - 1}}{{\bf{Y}}^{(0)}}
\end{equation}
where  $\textbf{Y}^\star$ is the probability of prediction of the target domain corresponding to different class labels,  $\boldsymbol{W}$ is an affinity matrix and $\boldsymbol{D}$ is the diagonal matrix.   


For sake of simplicity, we define $\alpha {\rm{ = }}\frac{1}{{1 + \mu }}$ and then Eq.(\ref{eq:Y_optimal}) is reformulated as Eq.(\ref{eq:Y_alpha_optimal}):

\begin{equation}\label{eq:Y_alpha_optimal}
{{\bf{Y}}^ \star } = {({\bf{D}} - \alpha {\bf{W}})^{ - 1}}{{\bf{Y}}^{(0)}}
\end{equation}

								
To sum up, at a given iteration, sub-problem (a) as in Eq.(\ref{eq:prob1}) searches a latent feature subspace $\textbf{Z}$ in closering both marginal and conditional data distributions between source and target while making use of source and current target labels in pushing away interclass data; sub-problem (b) as in Eq.(\ref{eq:prob2}) infers  through Eq.(\ref{eq:Y_alpha_optimal}) novel labels for target data  in line with source data labels while making use of the geometric structures of the underlying data manifolds in the current subspace $\textbf{Z}$. This iterative process eventually ends up in a latent feature subspace where : 1) the discrepancies of both marginal and conditional data distributions between source and target are narrowed; 2)  source and target data are rendered more discriminative thanks to the increase of interclass distances; and 3) the geometric structures of the underlying data manifolds are aligned.     

The complete learning algorithm  is summarized in Algorithm 1 - \textbf{DGA-DA}.

\begin{algorithm}[!h]
\caption{Discriminative Geometry Aware Domain Adaptation (\textbf{DGA-DA})}
	\KwIn{Data $\bf{X}$, Source domain labels ${\bf{Y}}_{\cal S}$, subspace dimension $k$, number of iterations $T$, regularization parameters $\lambda $ and $\alpha $}

\textbf{Step {1}}:  Initialize the iteration counter t=0 and compute ${{\bf{M}}_0}$ as in Eq.(\ref{eq:M0}).\\


\textbf{Step 2}: Initialize pseudo target labels ${{\bf{Y}}_{\cal T}}$ and projection space $\textbf{A}$:\\

(1) Solve the generalized eigendecomposition problem\cite{fortin2000augmented,long2013transfer} as in Eq.(\ref{eq:eig}) (replace ${{\bf{M}}_{{\bf{cyd}}}}$ by ${{\bf{M}}_0}$ ) and obtain adaptation matrix $\bf A$, then  embed data via the transformation, $\bf{Z} = {{\bf{A}}^T}{\bf{X}}$\;
(2)Initialize pseudo target labels ${{\bf{Y}}_{\cal T}}$ via a base classifier, \textit{e.g.}, 1-NN, based on source domain labels ${{\bf{Y}}_{\cal S}}$.

\While{not converged and $t<T$	}{
\textbf{Step 3}: Update projection space \textbf{A} \\
	
(i) Compute ${\bf{M}}_c$ (Eq.(\ref{eq:Mc})) 

(ii) Compute ${{\bf{M}}_{\hat c}} = {{\bf{M}}_{S \to T}} + {{\bf{M}}_{T \to S}}$ as in Eq.(\ref{eq:mttos}) and Eq.(\ref{eq:mstot}) via ${{\bf{Y}}_{\cal T}}$. 
		
(iii) Calculate ${{\bf{M}}_{cyd}} = {{\bf{M}}_c}+{{\bf{M}}_0} -{\bf M}_{\hat{c}} $;\\
(iv) Solve Eq.(\ref{eq:eig}) then update $\bf A$ and $\bf{Z} = {{\bf{A}}^T}{\bf{X}}$\;
										
\textbf{Step 4}: Label deduction \\
(i) construct the label matrix ${{\bf{Y}}^{(0)}}$ as in Eq.(\ref{eq:labelconsistency1})\;
(ii) design the affinity matrix\cite{NIPS2001_2092} $\boldsymbol{W}$ and diagonal matrix $\boldsymbol{D}$\;
(iii) obtain ${{\rm{{\bf{Y}}}}_{final}}$ in solving Eq.(\ref{eq:Y_optimal})\;

\textbf{Step  5}: update pseudo target labels $\{ {\bf{Y}}_{\cal T}^{(F)} = {{\bf{Y}}_{final}}\left[ {:,({n_s} + 1):({n_s} + {n_t})} \right]\} $;\\
\textbf{Step  6}: $t=t+1$; Return to Step 3; \\               
									
								}
\KwOut{Adaptation matrix ${\bf{A}}$, embedding ${\bf{Z}}$, Target domain labels ${\bf{Y}}_{\cal T}^{(F)}$}
		\end{algorithm}

\vspace{-10pt} 
			
	\subsection{Kernelization Analysis}
    \label{ssection:Kernelization Analysis}
	The proposed \textbf{DGA-DA} method can be extended to nonlinear problems in a Reproducing Kernel Hilbert Space via the kernel mapping $\phi :x \to \phi (x)$, or $\phi ({\bf{X}}):[\phi ({{\bf{x}}_1}),...,\phi ({{\bf{x}}_n})]$, and the kernel matrix ${\bf{K}} = \phi {({\bf{X}})^T}\phi ({\bf{X}}) \in {R^{n*n}}$. We utilize
	the Representer theorem to formulate Kernel \textbf{DGA-DA} as
	\begin{equation}\label{eq:kernel}
		\resizebox{0.9\hsize}{!}{%
			$\begin{array}{*{20}{l}}
			{\begin{array}{*{20}{l}}
				{\mathop {\min }\limits_{{{\bf{A}}^T}{\bf{KH}}{{\bf{K}}^T}{\bf{A}} = {\bf{I}}} \left( {\begin{array}{*{20}{l}}
						{\sum\limits_{c = 0}^C {tr({{\bf{A}}^T}{\bf{K}}{{\bf{M}}_c}{{\bf{K}}^T}{\bf{A}})}  + \lambda \left\| {\bf{A}} \right\|_F^2}
						{ + \sum\limits_{j = 1}^C {\sum\limits_{i = 1}^{{n_s} + {n_t}} {\left\| {{\bf{Y}}_{ij}^{(F)} - {\bf{Y}}_{ij}^{(0)}} \right\|} }  + {{\bf{Y}}^T}{\bf{LY}}}
						\end{array}} \right)}
				{ + \mathop {\max }\limits_{{{\bf{A}}^T}{\bf{KH}}{{\bf{K}}^T}{\bf{A}} = {\bf{I}}} tr({{\bf{A}}^T}{\bf{K}}{{\bf{M}}_{{\bf{\hat c}}}}{{\bf{K}}^T}{\bf{A}})}
				\end{array}}
			\end{array}$}
	\end{equation}

	\section{Experiments}
In this section, we verify and analyze in-depth the effectiveness of our proposed  domain adaptation model, \textit{i.e.}, \textbf{DGA-DA},  on 36 cross domain image classification tasks generated by permuting six datasets (see Fig.\ref{fig:data}). 
Sect.\ref{subsection:Benchmarks and Features} describes the benchmarks and the features. Sect.\ref{subsection:Baseline Methods} lists the baseline methods which the proposed \textbf{DGA-DA} is compared to. Sect.\ref{subsection: Experimental setup} presents the experimental setup and introduces in particular two partial DA methods, namely \textbf{CDDA} and \textbf{GA-DA}, in addition to the proposed \textbf{DGA-DA} based on our full DA model. Sect.\ref{subsection: Experimental Results and Discussion} discusses the experimental results in comparison with the state of the art. Sect.\ref{subsection: Convergence and Parameter Sensitivity} analyzes the convergence and parameter sensitivity of the proposed method. Sect.\ref{subsection:Analysis and Verification} further provides insight into the proposed DA model in visualizing the achieved feature subspaces through both synthetic and real data.

	\subsection{Benchmarks and Features}
    \label{subsection:Benchmarks and Features}
\begin{figure*}[h!]
	\centering
	\includegraphics[width=1\linewidth]{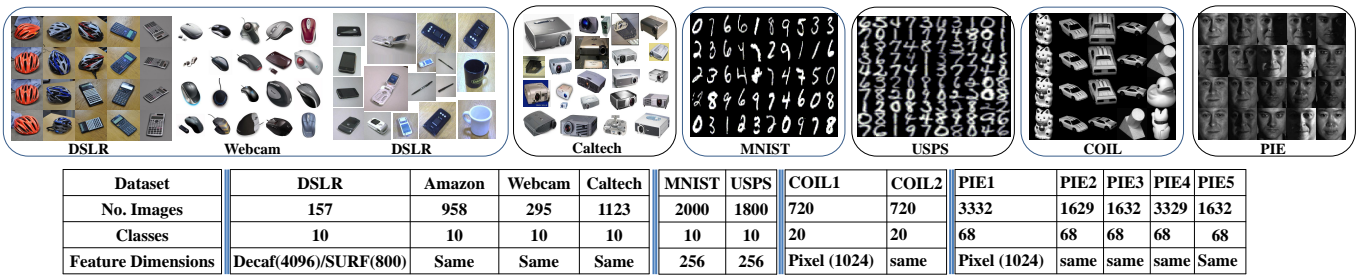}
	\caption { Sample images from six datasets used in our experiments. Each dataset represents a different domain. The OFFICE dataset contains three sub-datasets, namely DSLR, Amazon and Webcam.} 
	\label{fig:data}	
\end{figure*}

As illustrated in Fig.\ref{fig:data}, USPS\cite{DBLP:journals/pami/Hull94}+MINIST\cite{726791}, COIL20\cite{long2013transfer}, PIE\cite{long2013transfer} and office+Caltech\cite{long2013transfer,DBLP:journals/tip/XuFWLZ16,DBLP:journals/tip/HouTYW16,DBLP:journals/ijcv/ShaoKF14} are standard benchmarks for the purpose of evaluation and comparison with state-of-the-art in DA. In this paper, we follow the data preparation as most previous works\cite{DBLP:journals/tcyb/UzairM17,DBLP:journals/tip/XuFWLZ16,DBLP:conf/icml/GongGS13,DBLP:journals/pami/GhifaryBKZ17,DBLP:journals/tip/DingF17,DBLP:journals/corr/LuoWHC17} do. We construct 36 datasets for different image classification tasks.

\textbf{Office+Caltech} consists of 2533 images of ten categories (8 to 151 images per category per domain)\cite{DBLP:journals/pami/GhifaryBKZ17}. These images come from four domains: (A) AMAZON, (D) DSLR, (W) WEBCAM, and (C) CALTECH. AMAZON images were acquired in a controlled environment with studio lighting. DSLR consists of high resolution images captured by a digital SLR camera in a home environment under natural lighting. WEBCAM images were acquired in a similar environment to DSLR, but with a low-resolution webcam. CALTECH images were collected from Google Images. 

We use two types of image features extracted from these datasets, \textit{i.e}.,   \textbf{SURF} and \textbf{DeCAF6}, that are publicly available.  The \textbf{SURF}\cite{gong2012geodesic} features are \textit{shallow} features extracted and quantized into an 800-bin histogram using a codebook computed with K-means on a subset of images from  Amazon. The resultant histograms are further standardized by z-score. The \textbf{Deep Convolutional Activation Features (DeCAF6)}\cite{DBLP:conf/icml/DonahueJVHZTD14} are \textit{deep} features computed as in \textbf{AELM}\cite{DBLP:journals/tcyb/UzairM17} which makes of use VLFeat MatConvNet library with different pretrained CNN models, including in particular  the Caffe implementation of \textbf{AlexNet}\cite{krizhevsky2012imagenet} which is trained on the ImageNet dataset. The outputs from the 6th layer are used as \textit{deep} features, leading to 4096 dimensional \textbf{DeCAF6} features. In this experiment, we denote the dataset \textbf{Amazon},\textbf{Webcam},\textbf{DSLR},and \textbf{Caltech-256} as \textbf{A},\textbf{W},\textbf{D},and \textbf{C}, respectively. 

In denoting the direction from “source” to “target”  by an arrow “$\rightarrow$” is the direction from “source” to “target”, $4\times 3=12$ DA tasks can then be constructed, namely \emph{A} $\rightarrow$ \emph{W} $\dots$ \emph{C} $\rightarrow$ \emph{D}, respectively. For example, “W $\rightarrow$ D” means the Webcam image dataset is considered as the labeled \textit{source} domain whereas the DSLR image dataset the unlabeled \textit{target} domain. 
%
%


\textbf{USPS+MNIST} shares ten common digit categories from two subsets, namely USPS and MNIST, but with very different data distributions (see Fig.\ref{fig:data}). We construct a first DA task \emph{USPS vs MNIST} by randomly sampling 1,800 images in USPS to form the source data, and randomly sampling 2,000 images in MNIST to form the target data. Then, we switch the source/target pair to get another DA task, \textit{i.e.}, \emph{MNIST vs USPS}. We uniformly rescale all images to size 16×16, and represent each one by a feature vector encoding the gray-scale pixel values. Thus the source and target data share the same feature space. As a result, we have defined two cross-domain DA tasks, namely \emph{USPS $\rightarrow$ MNIST} and \emph{MNIST $\rightarrow$ USPS}.

\textbf{COIL20} contains 20 objects with 1440 images (Fig.\ref{fig:data}). The images of each object were taken in varying its pose about 5 degrees, resulting in 72 poses per object. Each image has a resolution of 32×32 pixels and 256 gray levels per pixel. In this experiment, we partition the dataset into two subsets, namely COIL 1 and COIL 2\cite{DBLP:journals/tip/XuFWLZ16}. COIL 1 contains all images taken within the directions in $[{0^0},{85^0}] \cup [{180^0},{265^0}]$ (quadrants 1 and 3), resulting in 720 images. COIL 2 contains all images taken in the directions
within $[{90^0},{175^0}] \cup [{270^0},{355^0}]$ (quadrants 2 and 4) and thus the number of images is 720. In this way, we construct two subsets with relatively different distributions. In this experiment, the COIL20 dataset with 20 classes is split into two DA tasks, \textit{i.e.},  \emph{ COIL1 $\rightarrow$ COIL2} and \emph{COIL2 $\rightarrow$ COIL1}

\textbf{PIE} face database consists of 68 subjects with each under 21 various illumination conditions\cite{DBLP:journals/tip/DingF17,long2013transfer}. We adopt five pose subsets: C05, C07, C09, C27, C29, which provides a rich basis for domain adaptation, that is, we can choose one pose as the source and any rest one as the target. Therefore, we obtain $5 \times 4=20$ different source/target combinations. Finally, we combine all five poses together to form a single dataset for large-scale transfer learning experiment. We crop all images to 32 × 32 and only adopt the pixel values as the input. Finally, with different face {poses}, of which five subsets are selected, denoted as PIE1, PIE2, \textit{etc}., resulting in $5 \times 4=20$ DA tasks, \textit{i.e.}, \emph{PIE1 vs PIE 2} $\dots$ \emph{PIE5 vs PIE 4}, respectively.

\subsection{Baseline Methods}
\label{subsection:Baseline Methods}
The proposed DGA-DA method is compared with \textbf{twenty-two} methods of the literature, including deep learning-based approaches for unsupervised domain adaption. They are: 
(1)1-Nearest Neighbor Classifier(\textbf{NN}); 
(2) Principal Component Analysis (\textbf{PCA}) +NN; 
(3) Geodesic Flow Kernel(\textbf{GFK}) \cite{gong2012geodesic} + NN; 
(4) Transfer Component Analysis(\textbf{TCA}) \cite{pan2011domain} +NN; 
(5) Transfer Subspace Learning(\textbf{TSL}) \cite{4967588} +NN; 
(6) Joint Domain Adaptation (\textbf{JDA}) \cite{long2013transfer} +NN; 
(7) Extreme Learning Machine (\textbf{ELM}) \cite{DBLP:journals/tcyb/UzairM17} +NN; 
(8) Augmented Extreme Learning Machine (\textbf{AELM}) \cite{DBLP:journals/tcyb/UzairM17} +NN; 
(9) Subspace Alignment (\textbf{SA})\cite{DBLP:conf/iccv/FernandoHST13}; 
(10) Marginalized Stacked Denoising Auto-encoder (\textbf{mSDA})\cite{DBLP:journals/corr/abs-1206-4683}; 
(11) Transfer Joint Matching (\textbf{TJM})\cite{DBLP:conf/cvpr/LongWDSY14}; 
(12) Robust Transfer Metric Learning (\textbf{RTML})\cite{DBLP:journals/tip/DingF17}; 
(13) Scatter Component Analysis (\textbf{SCA})\cite{DBLP:journals/pami/GhifaryBKZ17}; 
(14) Cross-Domain Metric Learning (\textbf{CDML})\cite{DBLP:conf/aaai/WangWZX14}; 
(15)Deep Domain Confusion (\textbf{DDC})\cite{DBLP:journals/corr/TzengHZSD14}; 
(16)Low-Rank Transfer Subspace Learning (\textbf{LTSL})\cite{DBLP:journals/ijcv/ShaoKF14}; 
(17)Low-Rank and Sparse Representation (\textbf{LRSR})\cite{DBLP:journals/tip/XuFWLZ16}; 
(18)Kernel Principal Component Analysis (\textbf{KPCA})\cite{DBLP:journals/neco/ScholkopfSM98}; 
(19)Joint geometric and statistical alignment (\textbf{JGSA}) \cite{Zhang_2017_CVPR}; 
(20)Deep Adaptation Networks (\textbf{DAN}) \cite{long2015learning};
(21)Deep Convolutional Neural Network (\textbf{AlexNet}) \cite{krizhevsky2012imagenet}
%
%
and
(22)Domain adaptation with low-rank reconstruction (\textbf{RVDLR}) \cite{jhuo2012robust}.

In addition, for the purpose of fair comparison, we follow   the experiment settings of \textbf{JGSA}, \textbf{AlexNet} and \textbf{SCA}, and apply DeCAF6 as the features for some methods to be evaluated. Whenever possible,  the reported performance scores of the \textbf{twenty-two} methods of the literature are directly  collected from previous research \cite{long2013transfer,DBLP:journals/tcyb/UzairM17,DBLP:journals/tip/DingF17,DBLP:journals/pami/GhifaryBKZ17,DBLP:journals/tip/XuFWLZ16,Zhang_2017_CVPR}. They are assumed to be their \emph{best} performance.

\subsection{Experimental Setup}
\label{subsection: Experimental setup}
For the problem of domain adaptation, it is not possible to tune a set of optimal hyper-parameters, given the fact that the target domain has no labeled data. Following the setting of previous research\cite{DBLP:journals/corr/LuoWHC17,long2013transfer,DBLP:journals/tip/XuFWLZ16} , we also evaluate the proposed \textbf{DGA-DA} by empirically searching in the parameter space for the \emph{optimal} settings. Specifically, the proposed \textbf{DGA-DA}  method has three hyper-parameters, \textit{i.e.}, the subspace dimension $k$, regularization parameters $\lambda $ and $\alpha $. In  our experiments, we set $k = 100$ and 1) $\lambda  = 0.1$, and $\alpha  = 0.99$ for \textbf{USPS}, \textbf{MNIST}， \textbf{COIL20} and \textbf{PIE}, 2) $\lambda  = 1$, $\alpha  = 0.99$ for \textbf{Office} and \textbf{Caltech-256}.
							
In our experiment, {\emph{accuracy}}  on the test dataset as defined by Eq.(\ref{eq:accuracy}) is the evaluation measurement. It is widely used in literature, \textit{e.g.},\cite{pan2008transfer,long2015learning,DBLP:journals/corr/LuoWHC17,long2013transfer,DBLP:journals/tip/XuFWLZ16}, \textit{etc}.
							
	\begin{equation}\label{eq:accuracy}
		\begin{array}{c}
	Accuracy = \frac{{\left| {x:x \in {D_T} \wedge \hat y(x) = y(x)} \right|}}{{\left| {x:x \in {D_T}} \right|}}
	\end{array}
	\end{equation}
where ${\cal{D_T}}$ is the target domain treated as test data, ${\hat{y}(x)}$ is the predicted label and ${y(x)}$ is the ground truth label for a test data  $x$.

To provide insight into the proposed DA method and highlight the individual contribution of each term in our final model, \textit{i.e.}, the discriminative term using the repulsive force as defined in Eq.(\ref{eq:repulsive}) and the geometry aware term through label smooth consistency as in Eq.(\ref{eq:labelconsitency}) and geometry structure consistency as in Eq.(\ref{eq:YLY}), we evaluate the proposed DA method using three settings:
\begin{itemize}
\item \textbf{CDDA}: In this setting,  sub-problem (b) in sect. \ref{subsection:solving the model} as defined in Eq.(\ref{eq:prob2})  is simply replaced by the Nearest Neighbor (NN) predictor.  This correspond to our final DA model as defined in Eq.(\ref{eq:prob1}) which only makes use of the \textit{repulse force} term but without geometry aware label inference as defined by Eq.(\ref{eq:labelconsitency}) and Eq.(\ref{eq:YLY}). This setting makes it possible to understand how important discriminative DA is \textit{w.r.t.} state of the art baseline DA methods only focused on data distribution alignment, \textit{e.g.}, \textbf{JDA}.

\item \textbf{GA-DA}: In this setting, we extend popular data distribution alignment-based DA methods,\textit{ e.g.}, \textbf{JDA}, with geometry aware label inference but ignore the \textit{repulsive force} term, \textit{i.e.},  $\max ({{\bf{A}}^T}{\bf{X}}{{\bf{M}}_{{\bf{\hat c}}}}{{\bf{X}}^T}{\bf{A}})$, in our final model reformulated in Eq.(\ref{eq:final model}). This setting thus jointly consider across domain conditional and marginal distribution alignment (Eq.(\ref{eq:M0}) and Eq.(\ref{eq:conditional})) and geometry aware label inference  (Eq.(\ref{eq:labelconsitency}) and Eq.(\ref{eq:YLY})). This setting enables  quantification of the contribution of the geometry aware label inference term as defined by Eq.(\ref{eq:labelconsitency}) and Eq.(\ref{eq:YLY}) in comparison with state of the art baseline DA methods only focused on data distribution alignment, \textit{e.g.}, \textbf{JDA}.

\item \textbf{DGA-DA}：This setting correspond to our full final model as defined in Eq.(\ref{eq:final model}). It thus contains \textbf{ CDDA} as expressed by  sub-problem (a) as in sect. \ref{subsection:solving the model} to which we further add the geometry aware label inference as defined by  sub-problem (b) in sect. \ref{subsection:solving the model}. 
\end{itemize}

\subsection{Experimental Results and Discussion}
\label{subsection: Experimental Results and Discussion}


\subsubsection{\textbf{Experiments on the COIL 20 Dataset}} 
\label{subsubsection: results on the COIL dataset}
\begin{figure}[h!]
		\centering
		\includegraphics[width=0.5\linewidth]{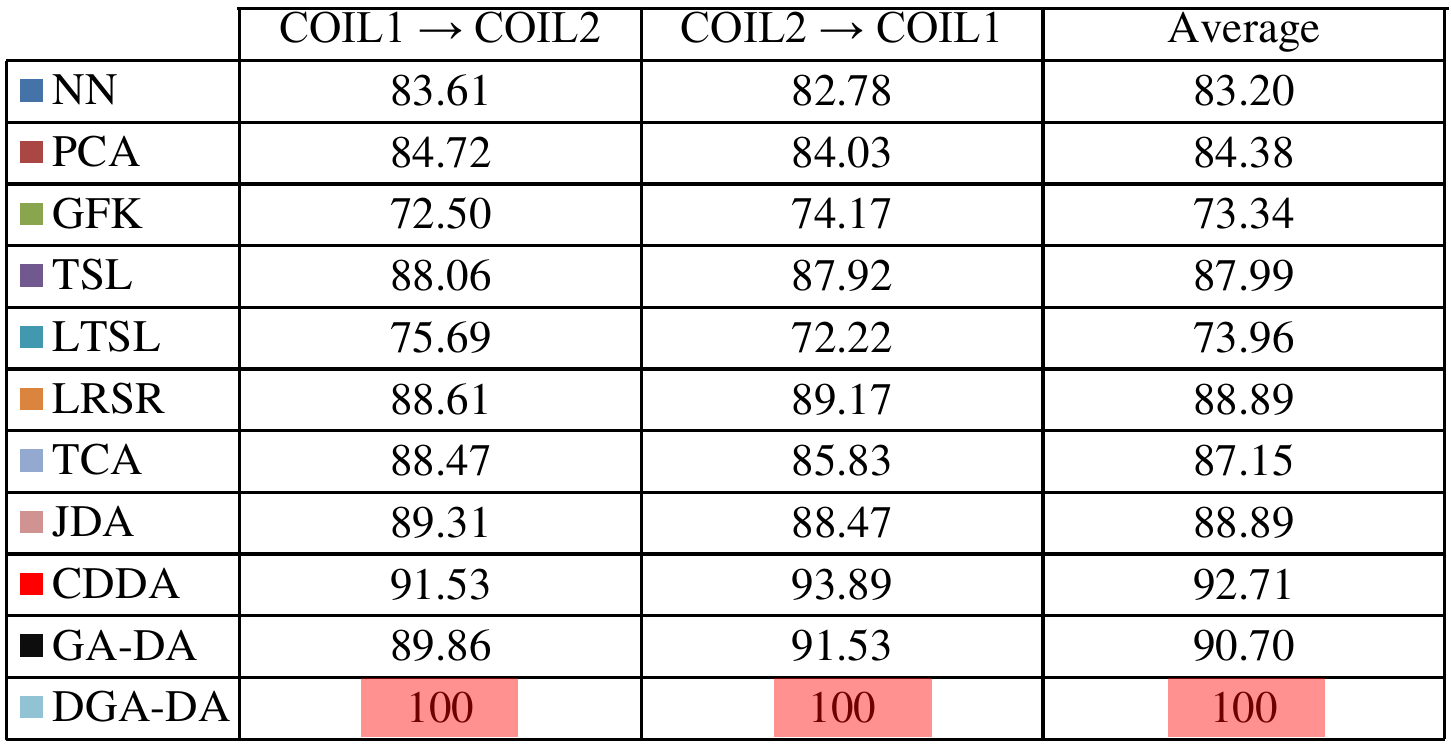}
		\caption { Accuracy${\rm{\% }}$ on the COIL Images Dataset.} 
		\label{fig:accCOIL}
	\end{figure}

The COIL dataset (see fig.\ref{fig:data}) features the challenge of pose variations between the source and target domain. Fig.\ref{fig:accCOIL} depicts the experimental results on the COIL dataset.  As can be seen in this figure where top results are highlighted in red color, the two partial models, \textit{i.e.},  \textbf{CDDA}, \textbf{DA-GA} and the proposed final model, \textbf{DGA-DA}, depict an overall average accuracy of  $\bf 92.71\%$, $\bf 90.70\%$ and $\bf 100.00\%$, respectively. They both outperform the eight baseline DA algorithms with a significant margin.

It is worth noting that, when adding label inference based on the underlying data manifold structure, the proposed \textbf{DGA-DA} improves its sibling \textbf{CDDA} by a margin as high as roughly 7 points, thereby highlighting the importance of data geometry aware label inference as introduced in \textbf{DGA-DA}. As compared to \textbf{JDA}, the proposed \textbf{CDDA}, which adds a discriminative \textit{repulsive force} term \textit{w.r.t.} \textbf{JDA}, also shows its effectiveness and improves the latter by more than 3 points. 




\vspace{3pt}
\subsubsection{\textbf{Experiments on the Office+Caltech-256 Data Sets}}
\label{subsubsection:Experiments on the Office+Caltech-256 Data Sets}

	\begin{figure*}[h!]
		\centering
		\includegraphics[width=1\linewidth]{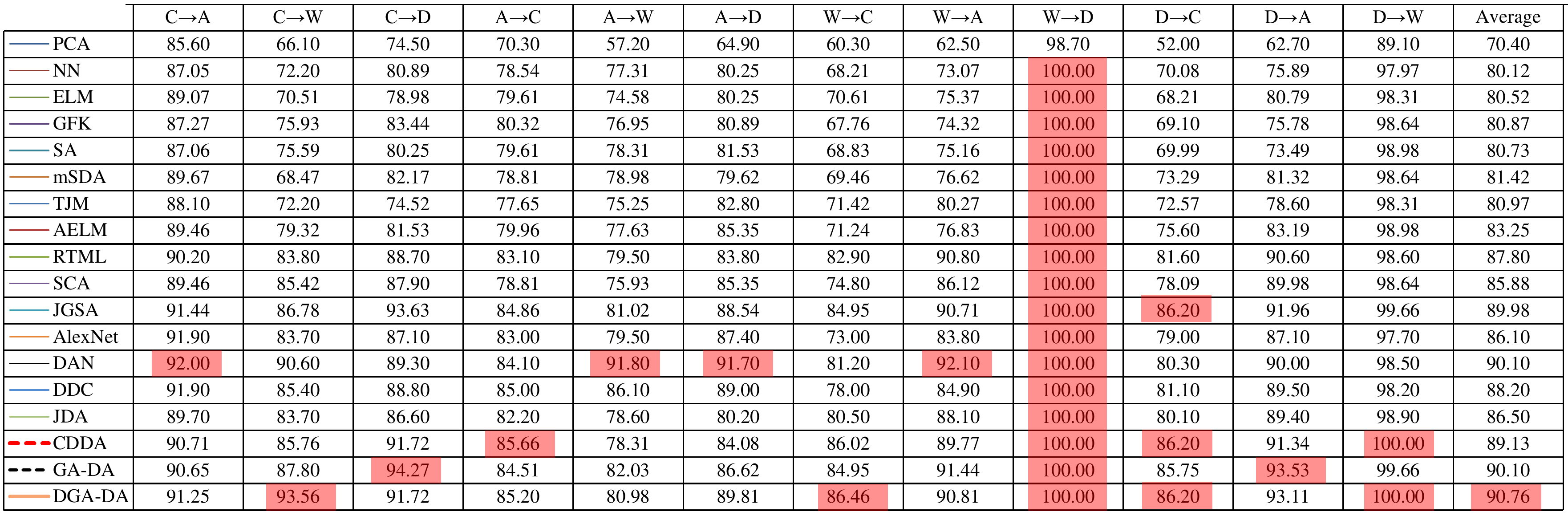}
		\caption { Accuracy${\rm{\% }}$ on the Office+Caltech Images with DeCAF6 Features.} 
		\label{fig:accDO}
	\end{figure*}

Fig.\ref{fig:accDO} and Fig.\ref{fig:accSO} synthesize the experimental results in comparison with the state of the art when deep features (\textit{i.e.}, DeCAF6 features) and classic shallow features (\textit{i.e.}, SURE features)   are used, respectively.


\begin{itemize}
\item  As can be seen in Fig.\ref{fig:accSO}, both \textbf{CDDA} and \textbf{DGA-DA} outperform the state of the art method in terms of average accuracy, thereby demonstrating the effectiveness of the proposed DA method. In particular, in comparison with \textbf{JDA} which only cares about data distribution alignment between source and target and the proposed DA method is built upon, \textbf{CDDA} improves \textbf{JDA} by 2 points thanks to the discriminative repulsive force term introduced in our model. When label inference accounts for the underlying data structure, our final model \textbf{DGA-DA} further improves \textbf{CDDA} by roughly 1 point.

	\begin{figure*}[h!]
		\centering
		\includegraphics[width=1\linewidth]{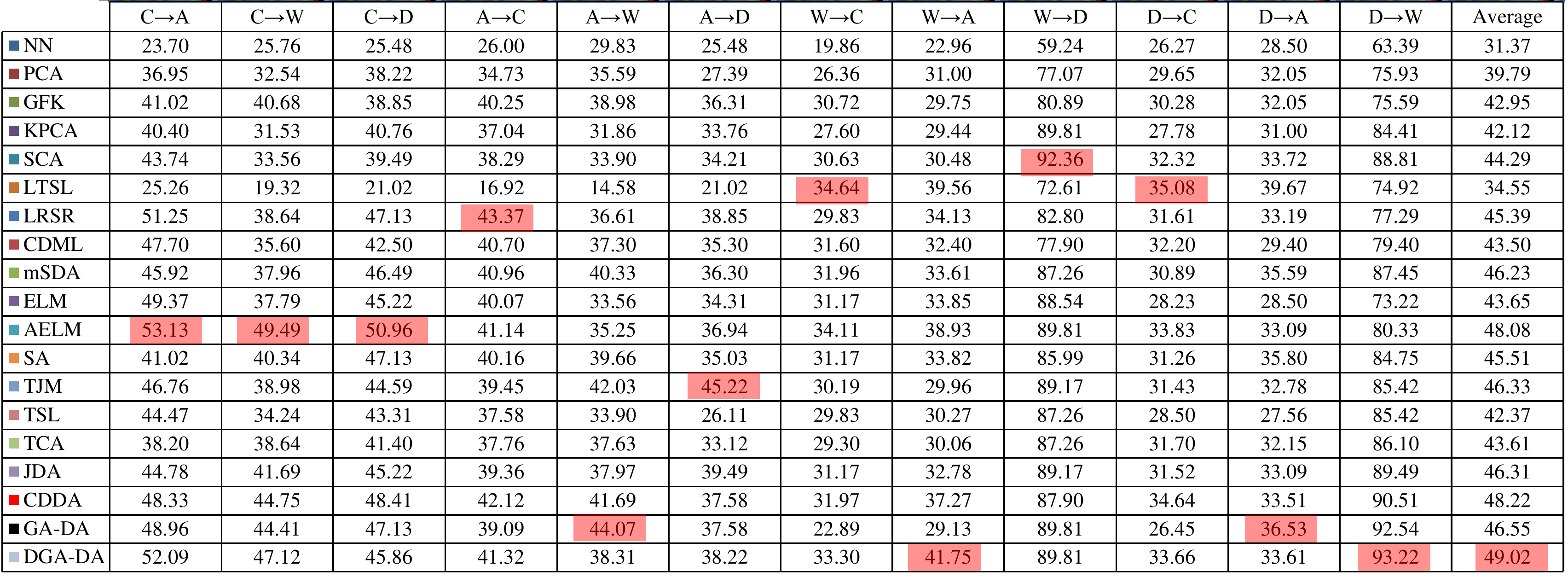}
		\caption { Accuracy${\rm{\% }}$ on the Office+Caltech Images with SURF-BoW Features.} 
		\label{fig:accSO}
	\end{figure*}

\item Fig.\ref{fig:accDO} compares the proposed DA method using deep features \textit{w.r.t.}   the state of the art, in particular end-to-end deep learning-based DA methods. As can be seen in Fig.\ref{fig:accDO}, the use of deep features has enabled impressive accuracy improvement over shallow features. Simple baseline methods, \textit{e.g.}, \textbf{NN}, \textbf{PCA}, see their accuracy soared by roughly 40 points, demonstrating the power of deep learning paradigm. Our proposed DA method also takes advantage of this jump and sees its accuracy soared from 48.22 to 89.13 for \textbf{CDDA} and from 49.02 to 90.43 for \textbf{DGA-DA}. As for shallow features, \textbf{CDDA} improves \textbf{JDA} by 3 points and \textbf{DGA-DA} further ameliorates \textbf{CDDA} by 1 point when label inference accounts for the underlying data geometric structure. As a result, \textbf{DGA-DA} displays the best average accuracy and outperforms slightly \textbf{DAN}.   

\end{itemize}

\subsubsection{\textbf{Experiments on the USPS+MNIST Data Set}}
\label{subsubsection: experiments on the UPS+MNIST Datasets}
\begin{figure}[h!]
		\centering
		\includegraphics[width=0.5\linewidth]{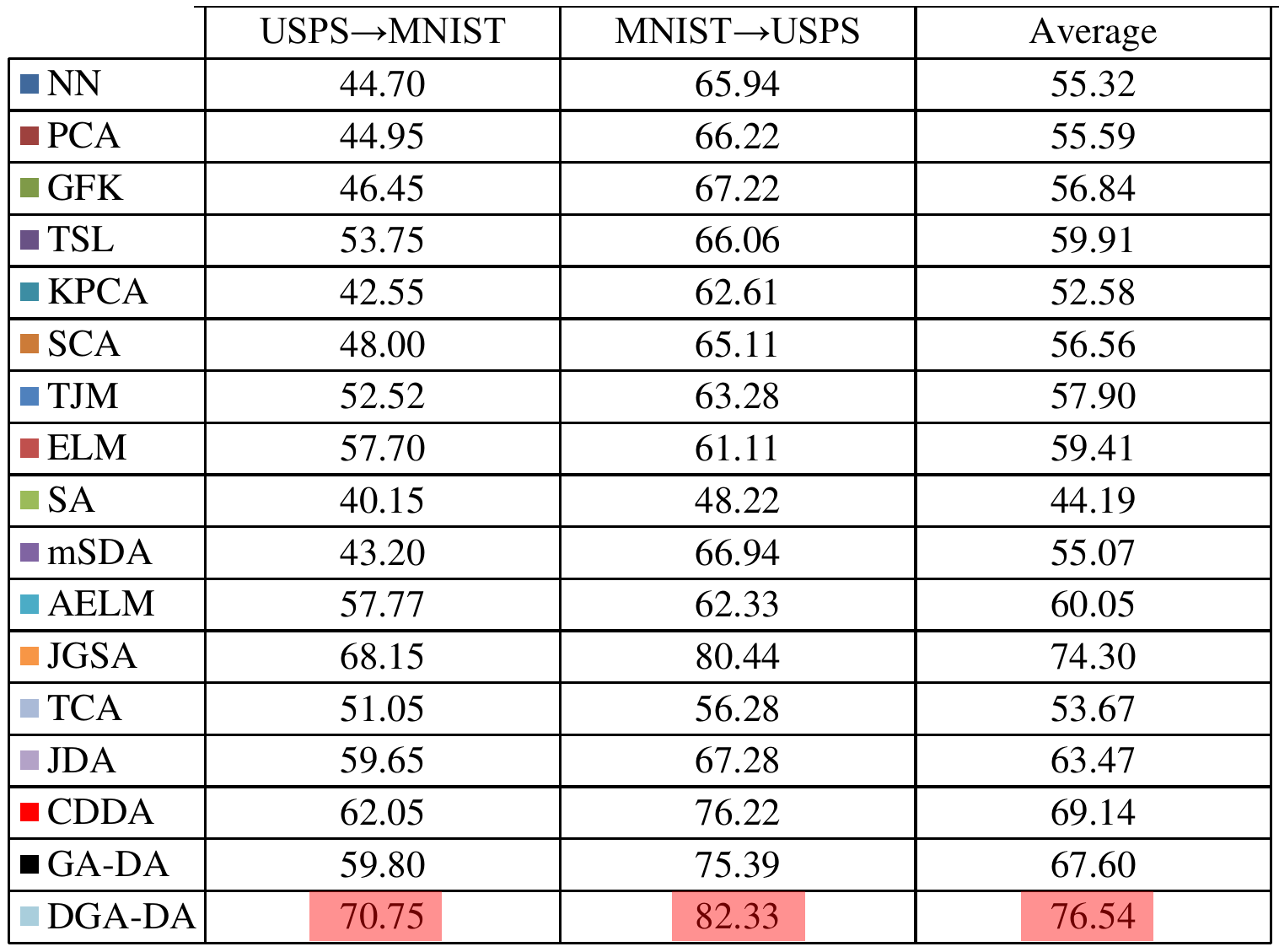}
		\caption { Accuracy${\rm{\% }}$ on the USPS+MNIST Images Dataset.} 
		\label{fig:accUSPS}
	\end{figure}

The UPS+MNIST dataset features different writing styles between source and target. Fig.\ref{fig:accUSPS} lists the experimental results in comparison with 14 state of the art DA methods. As can be seen in the table, \textbf{CDDA} displays a 69.14\% average accuracy and ranks the third best performer. It shows its effectiveness once more as it improves its baseline \textbf{JDA} by more than 5 points on average. When accounting for the underlying data geometry structure, the proposed \textbf{DGA-DA} further improves its sibling \textbf{CDDA} by a margin more than 7 points and displays the state of the art performance of a 76.54\% accuracy. It is worth noting that the second best DA performer on this dataset, \textit{i.e.}, \textbf{JGSA}, also suggests aligning both statistically and geometrically data, and thereby corroborates our data geometry aware DA approach.       

\subsubsection{\textbf{Experiments on the CMU PIE Data Set}}
\label{subsubsection:Experiments on the CMU PIE dataset}

\begin{figure*}[h!]
		\centering
		\includegraphics[width=1\linewidth]{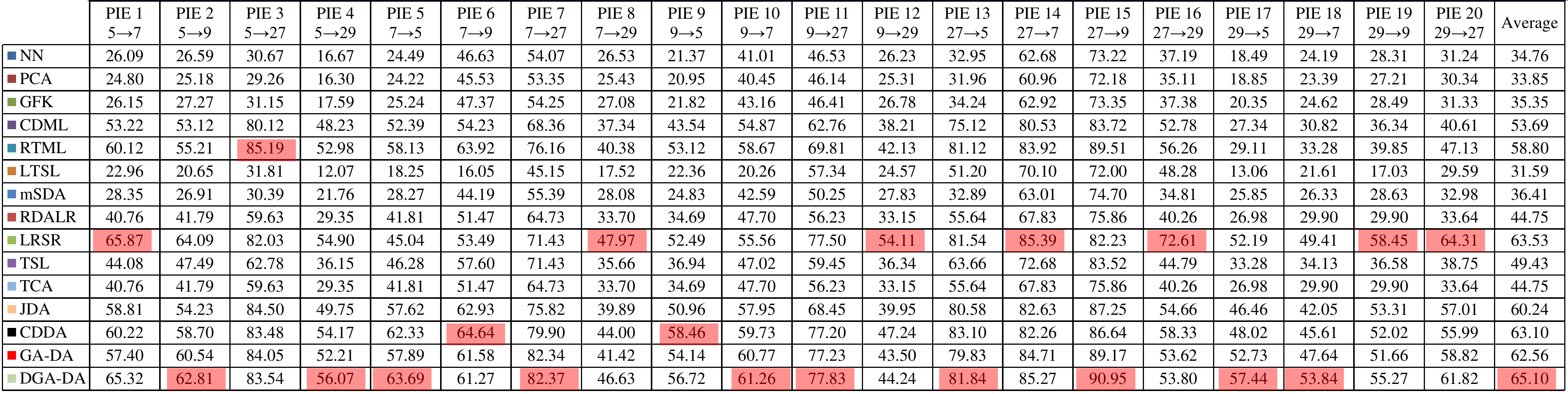}
		\caption { Accuracy${\rm{\% }}$ on the PIE Images Dataset.} 
		\label{fig:accPIE}
	\end{figure*}

The CMU PIE dataset is a large face dataset featuring both illumination and pose variations. Fig.\ref{fig:accPIE} synthesizes the experimental results for DA using this dataset. As can be seen in the figure, similarly as in the previous experiments, the proposed \textbf{DGA-DA} displays the best average accuracy over 20 cross-domain adaptation experiments. In aligning both marginal and conditional data distributions, \textbf{JDA} performs quite well and displays a 60.24\% average accuracy. In integrating the discriminative repulsive force term, \textbf{CDDA} improves \textbf{JDA} by roughly 3 points. \textbf{DGA-DA} further ameliorates \textbf{CDDA} by more than 1 point.   


It is interesting to note that the second best performer on this dataset, namely \textbf{LRSR}, also tries to align geometrically source and target data through both low rank and sparse constraints so that source and target data are interleaved within a novel shared feature subspace. 


\subsection{Convergence and Parameter Sensitivity }
\label{subsection: Convergence and Parameter Sensitivity}

\begin{figure*}[h!]
		\begin{center}
			\begin{tabular}{c}
	\includegraphics[width=1\linewidth]{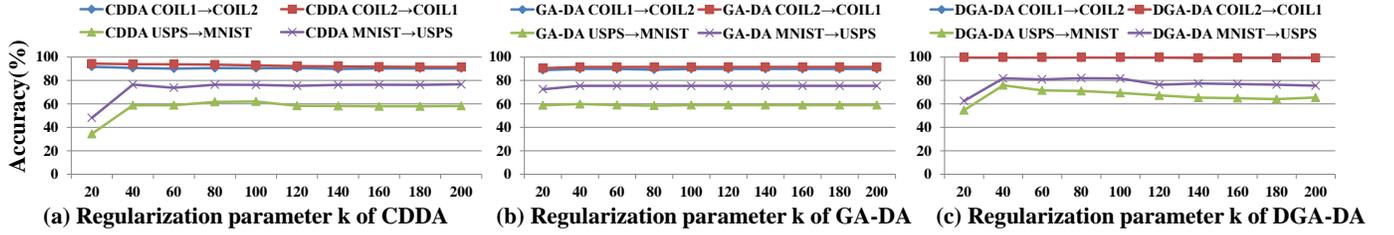}
			\end{tabular}
		\end{center}
        	\vspace{-10pt} 
		\caption {Sensitivity analysis of the proposed methods:  (a) accuracy \textit{w.r.t.} subspace dimension  $k$ of CDDA; (b)accuracy \textit{w.r.t.} subspace dimension $k$ of GA-DA;
(c) accuracy \textit{w.r.t.} subspace dimension $k$ of DGA-DA.
Four datasets are used, \textit{i.e.}, COIL1, COIL2, USPS and MNIST.} 
        		\label{fig:k}
	\end{figure*}

\begin{figure}[h!]
	\centering
	\includegraphics[width=0.5\linewidth]{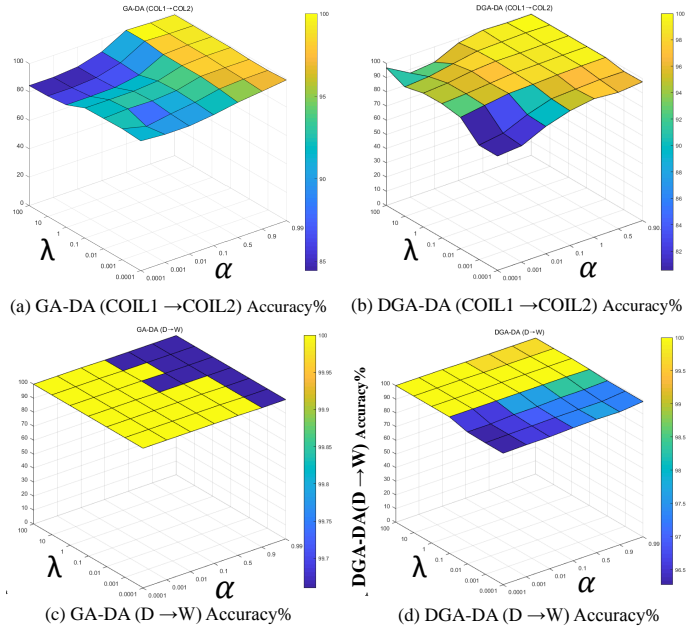}
	\caption {The classification accuracies of the proposed \textbf{GA-DA} and \textbf{DGA-DA} method vs. the parameters $\alpha $ and $\lambda $ on the selected four cross domains data sets, \textit{i.e.}, DSLR (D), Webcam (W), COIL1 and COIL2, with $k$ held fixed at $100$. } 
    	\label{fig:para}
\end{figure}

\begin{figure*}[h!]
	\centering
	\includegraphics[width=1\linewidth]{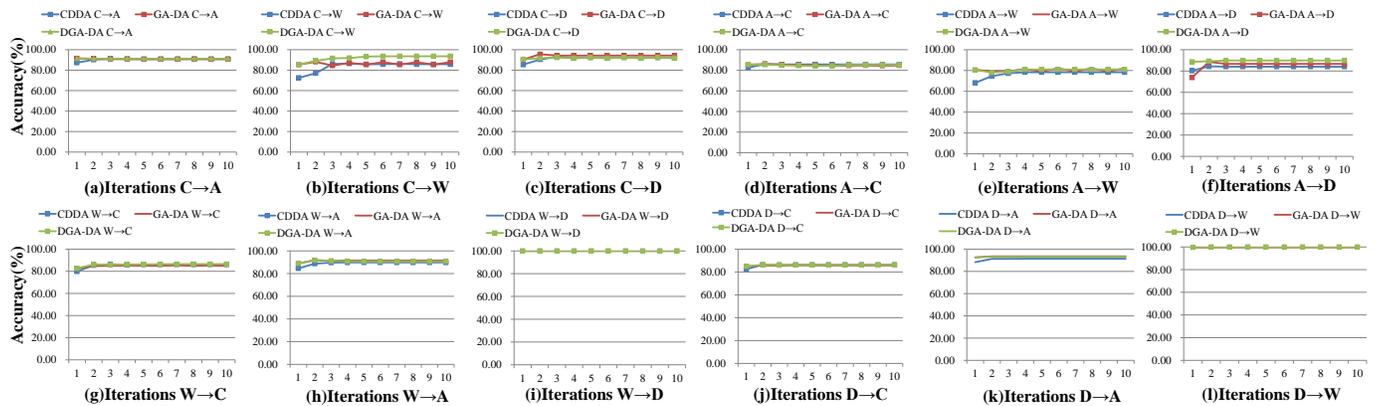}
	\caption {Convergence analysis using 12 cross-domain image classification tasks on Office+Caltech256 datasets with DeCAF6 Features. (accuracy w.r.t $\#$iterations) } 
    	\label{fig:accITER}
\end{figure*}

\begin{figure}[h!]
	\centering
	\includegraphics[width=1\linewidth]{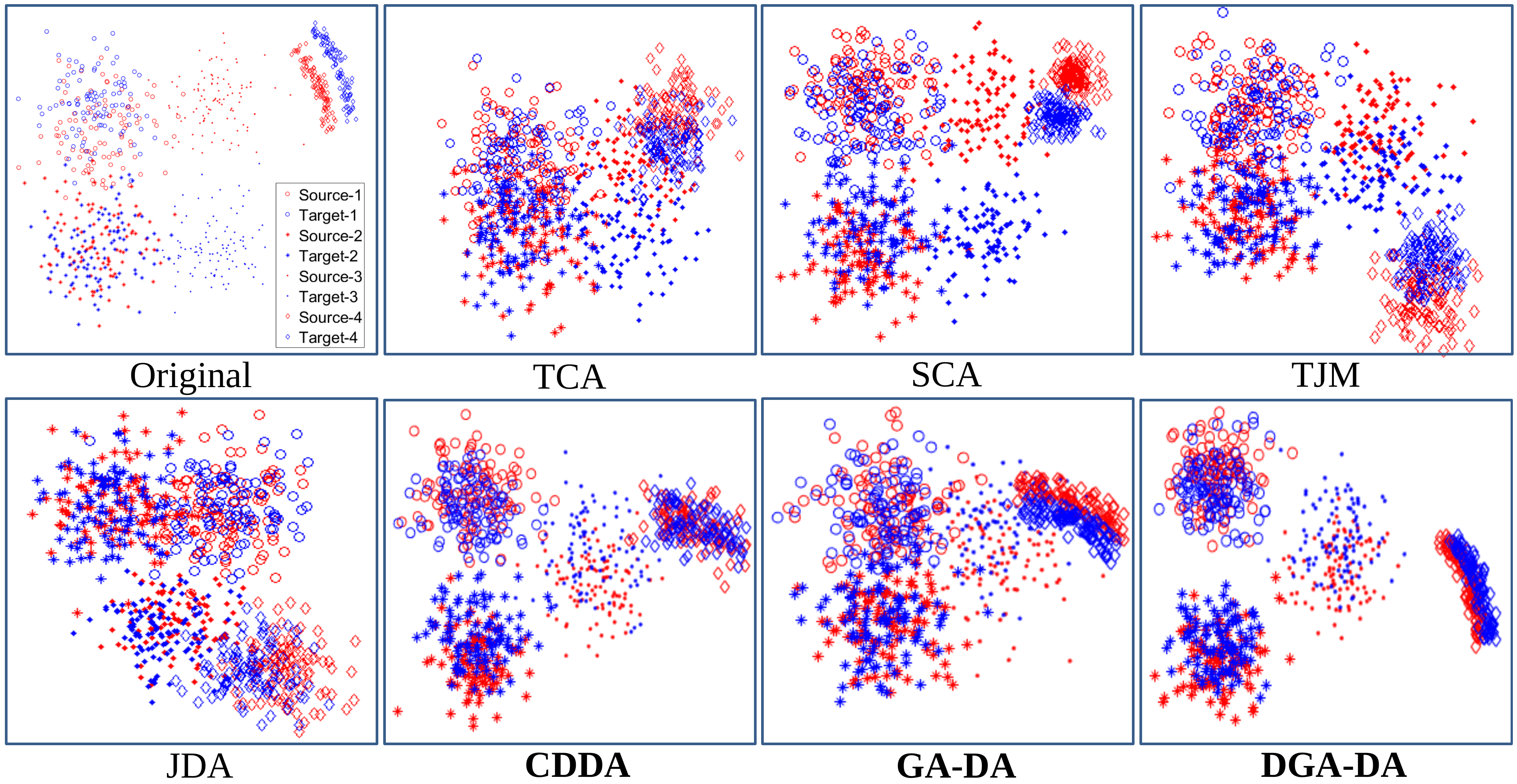}
	\caption {Comparisons of baseline domain adaptation methods and the proposed \textbf{CDDA}, \textbf{GA-DA} and \textbf{DGA-DA} method on the synthetic data} 
    	\label{fig:compare}
\end{figure}

\begin{figure*}[h!]
	\centering
	\includegraphics[width=1\linewidth]{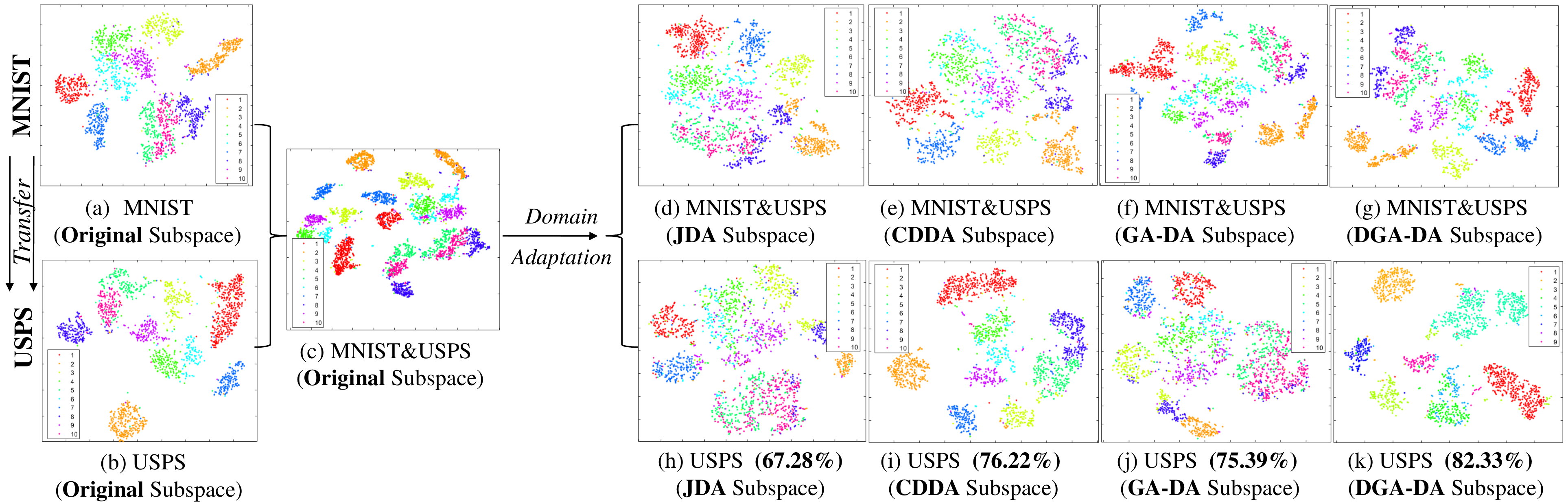}
	\caption {Accuracy(${\rm{\% }}$) and Visualization results of the MNIST$\rightarrow$USPS DA task. Fig.\ref{fig:vi}(a), Fig.\ref{fig:vi}(b) and Fig.\ref{fig:vi}(c) are visualization results of MNIST, USPS,  MNIST$\& $USPS datasets in their \textbf{Original} data space, respectively. After domain adaptation, Fig.\ref{fig:vi}(d), Fig.\ref{fig:vi}(e), Fig.\ref{fig:vi}(f) and Fig.\ref{fig:vi}(g) visualize the MNIST$\& $USPS datasets in \textbf{JDA}, \textbf{CDDA}, \textbf{GA-DA} and \textbf{DGA-DA} subspaces, respectively. Fig.\ref{fig:vi}(h), Fig.\ref{fig:vi}(i), Fig.\ref{fig:vi}(j) and Fig.\ref{fig:vi}(k) show the visualization results of the target domain USPS in \textbf{JDA}, \textbf{CDDA}, \textbf{GA-DA} and \textbf{DGA-DA} subspaces, respectively. The ten digit classes are represented by different colors.}  
    	\label{fig:vi}
\end{figure*}

While the proposed \textbf{DGA-DA} displays state of the art performance over 36 DA tasks through six datasets (USPS, MINIST, COIL20, PIE, Amazon, Caltech), an important question is how fast the proposed method converges (sect.\ref{subsubsection: Convergence analysis}) as well as its sensitivity \textit{w.r.t.} its hyper-parameters (Sect.\ref{subsubsection: Parameter sensitivity}). 

\subsubsection{Parameter sensitivity}
\label{subsubsection: Parameter sensitivity}
	
Three hyper-parameters, namely  $k$, $\lambda$ and $\alpha$, are introduced in the proposed methods.  $k$ is the dimension of the extracted feature subspace which determines the structure of low-dimension embedding. In Fig.\ref{fig:k}, we plot the classification accuracies of the proposed DA method \textit{w.r.t} different values of $k$ on the  \textbf{COIL} and \textbf{USPS+MINIST} datasets. As shown in Fig.\ref{fig:k}, the subspace dimensionality $k$ varies with $k  \in \{20,40,60,80,100,120,140,160,180,200\}$, yet the proposed 3 DA variants, namely, \textbf{CDDA}, \textbf{GA-DA} and \textbf{DGA-DA}, remain stable \textit{w.r.t.}  a wide range of with $k \in \{ 40 \le k  \le 200\} $. In our experiments,  we set $k = 100$ to balance efficiency and accuracy.

$\lambda$ as introduced in Eq.(\ref{eq:final model}) and Eq.(\ref{eq:prob1}) aims to regularize the projection matrix $\textbf{A}$ to avoid over-fitting the chosen shared feature subspace with respect to both source and target data.  $\alpha {\rm{ = }}\frac{1}{{1 + \mu }}$ as defined in Eq.(\ref{eq:Y_alpha_optimal}) is a trade-off parameter which balances  LSC and GSC. We study the sensitivity of the proposed \textbf{GA-DA} and \textbf{DGA-DA} methods with a wide range of parameter values, \textit{i.e.}, $\alpha  = (0.0001,0.001,0.01,0.1,0.5,0.9,0.99)$ and $\lambda  = (0.0001,0.001,0.01,0.1,1,10,100)$. We plot in Fig.\ref{fig:para} the results on \emph{D} $\rightarrow$ \emph{W} $ and $ \emph{COIL1} $\rightarrow$ \emph{COIL2} datasets on both methods with $k$ held fixed at $100$. As can be seen from Fig.\ref{fig:para}, the proposed \textbf{GA-DA} and \textbf{DGA-DA} display their stability as the resultant classification  accuracies remain roughly the same despite  a wide range of $\lambda $ and $\alpha $ values. 

\subsubsection{Convergence analysis}
\label{subsubsection: Convergence analysis}

In Fig.\ref{fig:accITER}, we further perform  convergence analysis of the proposed  \textbf{CDDA}, \textbf{GA-DA} and \textbf{DGA-DA} methods  using the \textbf{DeCAF6} features on the \textbf{Office+Caltech} datasets. The question here is how fast a DA method achieves its best performance \textit{w.r.t.} the number of iterations $T$.
Fig.\ref{fig:accITER} reports 12 cross domain adaptation experiments ( \emph{C} $\rightarrow$ \emph{A}, \emph{C} $\rightarrow$ \emph{W} ... \emph{D} $\rightarrow$ \emph{A} , \emph{D} $\rightarrow$ \emph{W}  ) with the number of iterations $T = (1,2,3,4,5,6,7,8,9,10)$. 
 
As shown in Fig.\ref{fig:accITER},  \textbf{CDDA}, \textbf{GA-DA} and \textbf{DGA-DA} converge within 3$ \sim $5 iterations during optimization.

\subsection{Analysis and Verification}
\label{subsection:Analysis and Verification}

To further gain insight of the proposed \textbf{CDDA}, \textbf{GA-DA} and \textbf{DGA-DA} \textit{w.r.t.} its domain adaptation skills, we also evaluate the proposed methods using a synthetic dataset in comparison with several state of the art DA methods. Fig.\ref{fig:compare} visualizes the original data distributions with 4 classes and the resultant shared feature subspaces as computed by \textbf{TCA}, \textbf{JDA}, \textbf{TJM}, \textbf{SCA}, \textbf{CDDA}, \textbf{GA-DA} and \textbf{DGA-DA}, respectively. In this experiment, we focus our attention on the ability of the DA methods to: :  (a) narrow the discrepancies of data distributions between source and target; (b) increase data discriminativeness; and (c) align data geometric structures between source and target. As such, the original synthetic data depicts slight distribution discrepancies between source and target for the first two class data, wide distribution mismatch for the third and fourth class data. Fourth class data further depict a moon like geometric structure.

As can be seen in Fig.\ref{fig:compare}, baseline methods, \textit{e.g.}, \textbf{TCA}, \textbf{SCA}, \textbf{TJM} have difficulties to align data distributions with wide discrepancies, \textit{e.g.}, third class data. \textbf{JDA} narrow data distribution discrepancies but lacks class data discriminativeness. The proposed variant \textbf{CDDA} ameliorates \textbf{JDA} and makes class data well separated thanks to the introduced \textit{repulsive force} term but falls short to preserve data geometric structure (see the fourth  moon like class data. The variant \textbf{GA-DA} align data distributions and preserves the underlying data geometric structures thanks to label smoothness consistency (LSC) and geometric structure consistency (GSC) but lacks data discriminativeness. In contrast, thanks to the joint consideration of data discriminativeness and geometric structure awareness, the proposed \textbf{DGA-DA} not only align data distributions compactly but also separate class data very distinctively. Furthermore, it also preserves the underlying data geometric structures.


The above findings can be further verified using real data through the MNIST$\rightarrow$USPS DA task  where the proposed DA methods achieves remarkable results (See Fig.\ref{fig:accUSPS}). Fig.\ref{fig:vi} visualizes class explicit data distributions in their original subspace and  the resultant shared feature subspace using \textbf{JDA} and the three variants of the proposed DA method, namely \textbf{CDDA}, \textbf{GA-DA} , \textbf{DGA-DA}, with the same experimental setting. 



\begin{itemize}
	
	\item Data distributions and geometric structures. Fig.\ref{fig:vi}(a,b,c) visualize the MNIST, USPS,  MNIST$\& $USPS datasets in their \textbf{Original} data space, respectively. As shown in these figures, the MNIST and USPS datasets depict different data distributions and various data structures. In particular, yellow  dots represent digit 2. They show a long and narrow shape in MNIST (Fig.\ref{fig:vi}(a)) while a circle like shape in USPS (Fig.\ref{fig:vi}(b)). They further display large data discrepancies across domain (Fig.\ref{fig:vi}(c)) as for all the other classes.

    \item Contribution of the \textit{repulsive force} term. Visualization results in Fig.\ref{fig:vi}(h,i,j,k) show that, in comparison with their respective baseline DA methods, \textit{i.e.}, \textbf{JDA} (Fig.\ref{fig:vi}(h)) and \textbf{GA-DA} (Fig.\ref{fig:vi}(j)), the proposed two DA variants, \textit{i.e.}, \textbf{CDDA}(Fig.\ref{fig:vi}(i))  and \textbf{DGA-DA}(Fig.\ref{fig:vi}(k)) which integrate in their model the \textit{repulsive force} term as introduced in Sect.\ref{subsubsection:Discriminative DA}, achieve data discriminativeness in compacting intra-class instances and separating inter-class data, respectively. As a result,  as shown in Fig.\ref{fig:accUSPS}, \textbf{DGA-DA} outperforms \textbf{GA-DA} by ${\bf 6.94}\uparrow$ points, and \textbf{CDDA} outperforms \textbf{JDA} by ${\bf 8.94}\uparrow$ points,respectively, thereby illustrating the importance of increasing data discriminativeness in DA.

    \item Contribution of Geometric Structure Awareness. Visualization results in Fig.\ref{fig:vi}(d,e) show that the \textbf{JDA} and \textbf{CDDA}'s subspaces fail to preserve the geometric structures of the underlying data manifold. For instance, the long and narrow shape of the orange dots in the source MNIST domain and the corresponding circle blob orange cloud in the target USPS domain (Fig.\ref{fig:vi}(c)) are not preserved anymore in the \textbf{JDA} (Fig.\ref{fig:vi}(d)) and \textbf{CDDA} (Fig.\ref{fig:vi}(e)) subspaces. In contrast, thanks to the geometry awareness constraints, \textit{i.e.}, label smoothness consistency (LSC) and geometric structure consistency (GSC), as introduced in Sect.\ref{subsubsection:GA-DA}, the two variants of the proposed DA methods, \textit{i.e.}, \textbf{DA-GA} (Fig.\ref{fig:vi}(f)) and \textbf{DGA-DA} (Fig.\ref{fig:vi}(g)), succeed to preserve the geometric structures of the underlying data, and thereby inherent data similarities and consistencies of label inference. As a result,  \textbf{DGA-DA} outperforms \textbf{CDDA} by ${\bf 6.11}\uparrow$ points, and \textbf{GA-DA} outperforms \textbf{JDA} by ${\bf 8.11}\uparrow$ points. They thus suggest the importance of Geometric Structure Awareness in DA. 

\end{itemize}



\section{Conclusion and Future Work}
In this paper, we have proposed a novel Discriminative and Geometry Aware Unsupervised DA method based on feature adaptation. Comprehensive experiments on 36 cross-domain image classification tasks through six popular DA datasets highlight the interest of enhancing the data discriminative properties within the model and label propagation in respect of the geometric structure of the underlying data manifold, and  verify the effectiveness of the proposed method compared with twenty-two baseline DA methods of the literature. Using both synthetic and real data and three variants of the proposed DA method, we have further provided in-depth analysis and insights into the proposed \textbf{DGA-DA}, in quantifying and visualizing the contribution of the data discriminativeness and data geometry awareness.

Our future work will concentrate on embedding the proposed method in  deep  networks and study other vision tasks, \textit{e.g.}, object detection, within the setting of transfer learning.
Our future work will concentrate on embedding the proposed method in  deep  networks and study other vision tasks, \textit{e.g.}, object detection, within the setting of transfer learning.

\ifCLASSOPTIONcaptionsoff
  \newpage
\fi



%

\small\bibliographystyle{plain}
\small\bibliography{cdda}

%
%

	\vspace{-10pt} 
\begin{IEEEbiography}   [{\includegraphics[width=1in,height=1.25in,clip,keepaspectratio]{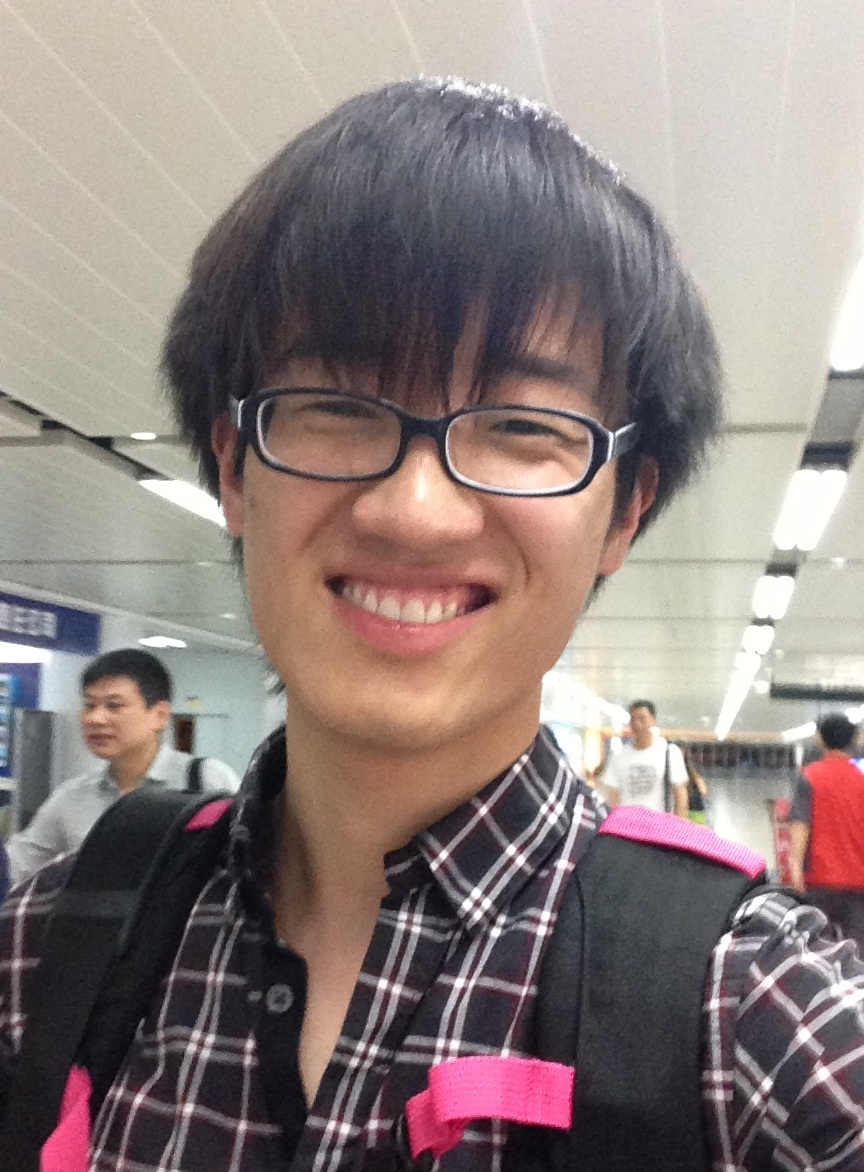}}]{Lingkun Luo}
	received his first master's degree in computer science from Hosei University and second master's degree in software engineering from University of Science and Technology of China. Now, he is a PhD candidate at Shanghai Jiao Tong University (SJTU). He is currently a research assistant in Ecole Centrale de Lyon (ECL), Department of Mathematics and Computer Science, and a member of LIRIS laboratory. He is jointly supervised by SJTU and ECL. His research interests include machine learning, pattern recognition and computer vision.
\end{IEEEbiography}
	\vspace{-5pt} 
\begin{IEEEbiography}[{\includegraphics[width=1in,height=1.25in,clip,keepaspectratio]{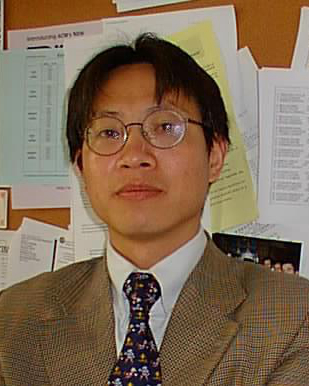}}]{Liming Chen} received the joint B.Sc. degree in mathematics and computer science from the University of Nantes, Nantes, France in 1984, and the M.Sc. and Ph.D. degrees in computer science from the University of Paris 6, Paris, France, in 1986 and 1989, respectively.

He first served as an Associate Professor with the Universit\'{e} de Technologie de Compi\`{e}gne, before joining \'Ecole Centrale de Lyon, \'Ecully, France, as a Professor in 1998,  where he leads an advanced research team on multimedia computing and pattern recognition. From 2001 to 2003, he also served as Chief Scientific Officer in a Paris-based company, Avivias, specializing in media asset management. In 2005, he served as Scientific Multimedia Expert for France Telecom R\&D China, Beijing, China. He was the Head of the Department of Mathematics and Computer Science, \'Ecole Centrale de Lyon from 2007 through 2016. His current research interests include computer vision, machine learning, image and video analysis and categorization, face analysis and recognition, and affective computing. Liming has over 250 publications and successfully supervised over 35 PhD students. He has been a grant holder for a number of research grants from EU FP program, French research funding bodies and local government departments. Liming has so far guest-edited 3 journal special issues. He is an associate editor for Eurasip Journal on Image and Video Processing and a senior IEEE member.
\end{IEEEbiography}
	\vspace{-5pt} 
\begin{IEEEbiography}   [{\includegraphics[width=1in,height=1.25in,clip,keepaspectratio]{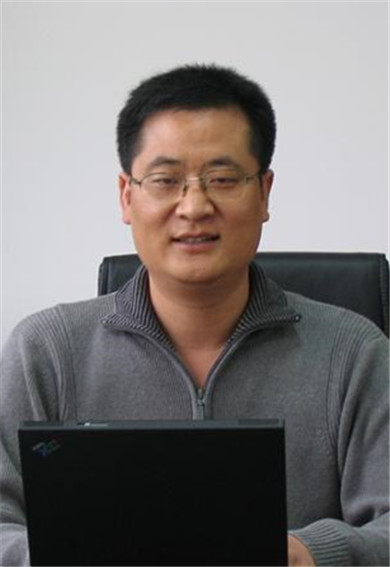}}]{Shiqiang Hu}
	received his PhD degree at Beijing Institute of Technology. His research interests include data fusion technology, image understanding, and nonlinear filter.
\end{IEEEbiography}
	\vspace{-5pt} 
\begin{IEEEbiography}   [{\includegraphics[width=1in,height=1.25in,clip,keepaspectratio]{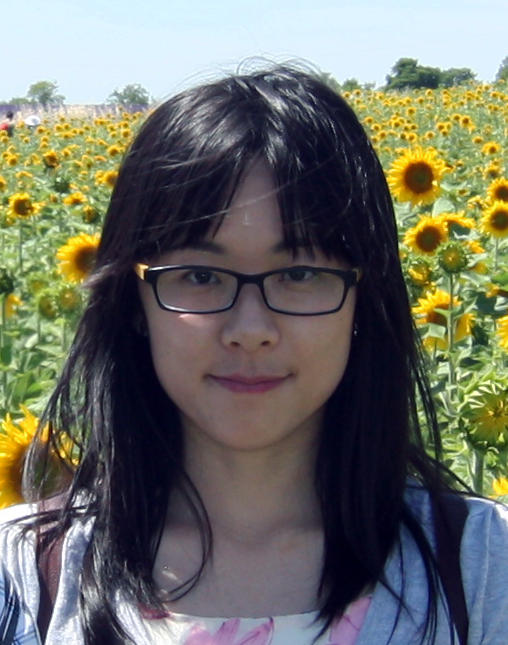}}]{Ying LU}
received the B.S. degree in Applied Mathematics and the M.S. degree in Computer Science and Engineering from Beihang University, Beijing, China, in 2010 and 2013, respectively, and Ph.D. degree in computer science from University of Lyon, France, in 2017. She also received a Research Master's degree in Computer Science from Ecole Centrale de Lyon, University Lyon I and INSA Lyon in 2012, and an Engineering degree from Ecole Centrale de Pékin, Beihang Universy in 2013. She is currently a teaching and research assistant in Ecole Centrale de Lyon, Department of Mathematics and Computer Science, and a member of LIRIS laboratory. Her research interests include machine learning and computer vision.

\end{IEEEbiography}
	\vspace{-5pt} 
\begin{IEEEbiography}[{\includegraphics[width=1in,height=1.25in,clip,keepaspectratio]{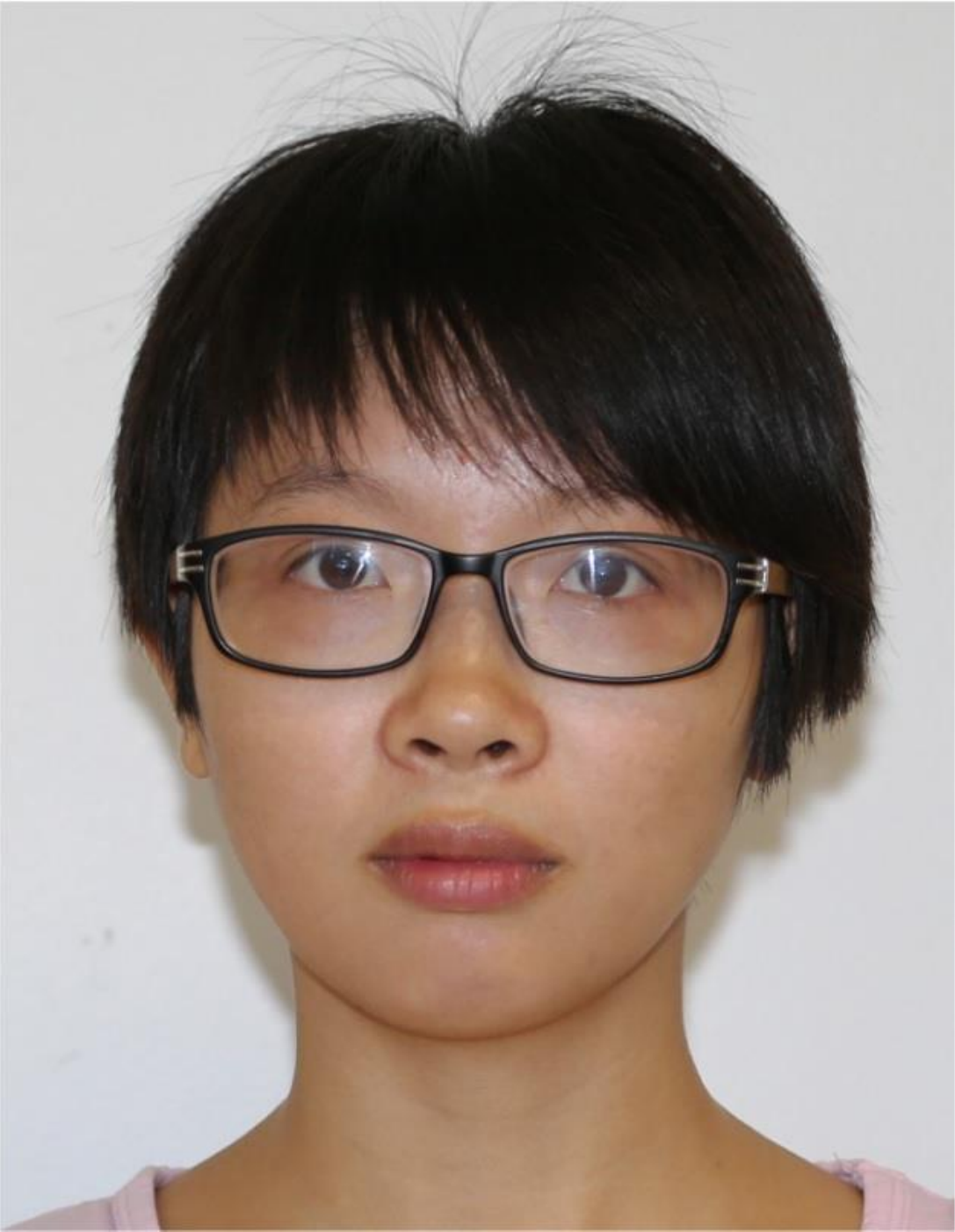}}]{Xiaofang Wang} is currently assistant lecturer and researcher in Ecole Centrale Lyon. She has received the B.S. and M.S. degrees in biomedical engineering from Central South University, Changsha, China, and the Ph.D. degree in computer science from \'Ecole Centrale de Lyon, France in 2015.

Her current research interests include image/video processing, machine learning (transfer learning, deep learning), computer vision (semantic image segmentation, object localization and recognition, etc.).
\end{IEEEbiography}

\end{CJK*}
\end{document}